\documentclass{article}

\PassOptionsToPackage{numbers}{natbib}

\usepackage[final]{neurips_2025}

\usepackage[utf8]{inputenc} 
\usepackage[T1]{fontenc}    
\usepackage{url}            
\usepackage{booktabs}       
\usepackage{amsfonts, amssymb, amsmath}       
\usepackage{mathtools}
\usepackage{dsfont}
\usepackage{nicefrac}       
\usepackage{microtype}      
\usepackage{array}
\usepackage{rotating}
\usepackage{pifont}
\usepackage{import}
\usepackage{physics}
\usepackage{bm, bbm}
\usepackage{multirow}
\usepackage{subcaption}
\usepackage[toc,page]{appendix}
\usepackage{tablefootnote}
\usepackage{wrapfig}
\usepackage{xspace}
\usepackage{enumitem}
\usepackage{blindtext}
\usepackage{algorithm}
\usepackage{tabularx}
\usepackage{float}
\usepackage{algpseudocode}
\usepackage{graphicx}
\usepackage[table]{xcolor}
\usepackage{pifont}
\usepackage{bbm}

\makeatletter
\newcommand{\multiline}[1]{%
    \begin{tabularx}{\dimexpr\linewidth-\ALG@thistlm}[t]{@{}X@{}}
        #1
    \end{tabularx}
}

\definecolor{gold}{RGB}{255, 215, 0}
\definecolor{silver}{RGB}{192, 192, 192}
\definecolor{bronze}{RGB}{205, 127, 50}

\definecolor{darkpastelblue}{rgb}{0.47, 0.62, 0.8}
\usepackage[hidelinks,colorlinks=true,linkcolor=purple,citecolor=darkpastelblue]{hyperref}       

\usepackage[capitalize]{cleveref}

\crefname{section}{Sec.}{Secs.}
\crefname{equation}{Eq.}{Eqs.}
\crefname{figure}{Fig.}{Figs.}
\crefname{table}{Tab.}{Tabs.}
\crefname{appendix}{Supp.}{Supp.}

\newcommand{\cmark}{\ding{51}}
\newcommand{\xmark}{\ding{55}}

\newcolumntype{M}[1]{>{\centering\arraybackslash}m{#1}}
\newcolumntype{T}[1]{>{\small\centering\arraybackslash}m{#1}}

\title{JAFAR: Jack up Any Feature at Any Resolution}

\author{%
$^*$Paul Couairon$^{1,2}$
   \quad $^*$Loïck Chambon$^{1,3}$ \quad Louis Serrano$^{1}$ \\
   \quad \textbf{Jean-Emmanuel Haugeard}$^{2}$ \quad \textbf{Matthieu Cord}$^{1, 3}$ \quad \textbf{Nicolas Thome}$^{1, 4}$ \\\textnormal{$^1$Sorbonne Université, CNRS, ISIR, F-75005 Paris, France \, $^2$Thales, TSGF, cortAIx Labs, France} \\ \textnormal{$^3$Valeo.ai} \, $^4$Institut Universitaire de France (IUF) 
}

\begin{document}

\maketitle

\begin{figure}[!ht]
\makebox[1.\textwidth]{
  \centering
  \includegraphics[scale=0.6]{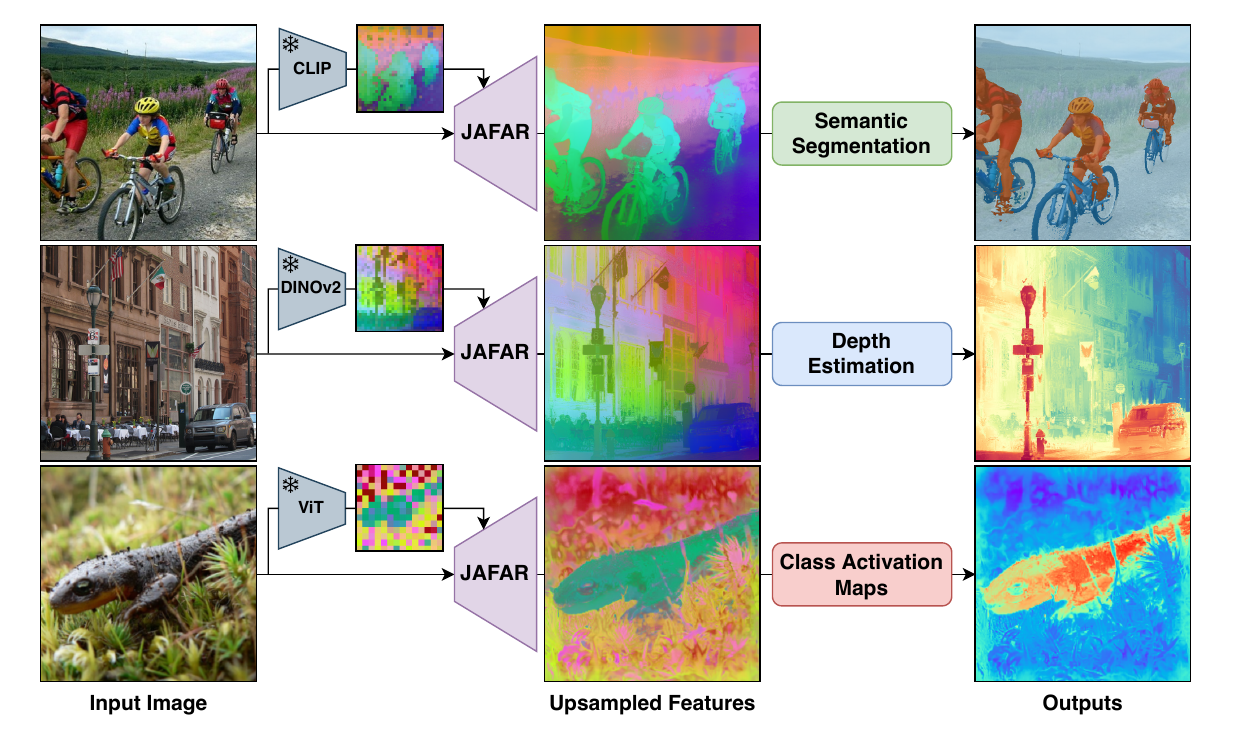}}
  \caption{JAFAR upsamples features from any foundation vision encoder to any image resolution, using the input image as high-resolution guidance. It generates sharp, boundary-aligned feature maps and serves as a versatile drop-in module for a variety of downstream tasks, including semantic segmentation, open-vocabulary segmentation, depth estimation, CAM evaluation, and bird’s-eye-view segmentation—consistently enhancing performance.}
  \label{fig:teaser}
\end{figure}

\begin{abstract}

Foundation Vision Encoders have become essential for a wide range of dense vision tasks. However, their low-resolution spatial feature outputs necessitate feature upsampling to produce the high-resolution modalities required for downstream tasks. In this work, we introduce JAFAR, a lightweight and flexible feature upsampler that enhances the spatial resolution of visual features from any Foundation Vision Encoder to an arbitrary target resolution. JAFAR employs an attention-based module designed to promote semantic alignment between high-resolution queries, derived from low-level image features, and semantically enriched low-resolution keys, using Spatial Feature Transform (SFT) modulation. Notably, despite the absence of high-resolution supervision, we demonstrate that learning at low upsampling ratios and resolutions generalizes remarkably well to significantly higher output scales. Extensive experiments show that JAFAR effectively recovers fine-grained spatial details and consistently outperforms existing feature upsampling methods across a diverse set of downstream tasks.

\textbf{Project page}: \href{https://jafar-upsampler.github.io}{https://jafar-upsampler.github.io} 

\end{abstract}
\section{Introduction}
\label{section:introduction}

Foundation vision encoders—whether trained with language supervision \cite{clip, siglip, siglip2, tulip, radiov2.5} or purely on visual data \cite{dino, dinov2, capi}—have become core components of modern computer vision pipelines. Vision-language models excel at tasks requiring generalization, such as zero-shot classification and open-vocabulary segmentation \cite{maskclip, sclip}. In contrast, image-only models, which focus on visual structure, often outperform in dense prediction tasks that demand fine-grained spatial reasoning, including semantic segmentation, depth estimation, object discovery, and point tracking \cite{depth_anything, dino_meets_text, dinotracker}.

To handle high-resolution inputs and large-scale training, foundation vision encoders typically downsample spatial information aggressively—by a factor of $14 \times$ to $16 \times$—yielding semantically rich but spatially coarse feature maps. This compression introduces a bottleneck for downstream tasks that require pixel-level accuracy. As a result, downstream pipelines \cite{detr, odise, depth_anything, depth_anythingv2, diffcut} often rely on interpolation or dedicated modules \cite{dpt, mask2former} designed to produce high-resolution outputs.

Several strategies have been explored to overcome this bottleneck, but each comes with trade-offs in efficiency and output quality. A straightforward solution is to apply training-free interpolation methods, such as bilinear upsampling. While computationally efficient, these direct interpolations—relying solely on low-resolution feature maps—fail to leverage information from the original high-resolution image, often resulting in blurry outputs. Alternatively, one can upsample the input image prior to encoding to increase feature resolution. However, this approach significantly increases computational cost due to the quadratic complexity of self-attention—common in foundation models—and may introduce artifacts in the feature maps, ultimately degrading performance \cite{dvt, vitar}.

Focusing specifically on a target downstream task, \cite{carafe, sapa, dysample, resfu, FADE} learn feature upsamplers using high-resolution supervision from task-specific labels. While generally lightweight, these upsamplers depend on labeled data tied to the end application, which limits their generalization and may bias the learned features toward optimizing task-specific losses. To address this, recent methods such as LiFT~\cite{lift} and FeatUp~\cite{featup} adopt task-agnostic training objectives. LiFT is trained to perform $2\times$ upsampling by regressing feature maps extracted from images at twice the input resolution. However, its convolution-based architecture is limited to fixed $2 \times$  scaling, restricting its flexibility for arbitrary output resolutions. FeatUp, in contrast, uses augmented views and self-reconstruction to support higher upsampling ratios. Yet, its Joint Bilateral Upsampling (JBU) variant suffers from over-smoothed outputs, while its implicit variant requires training the upsampler for each image, making it impractical in real-world scenarios.

In this paper, we introduce a feature upsampler designed to satisfy the following criteria: \textit{(i)} a task-agnostic training objective, \textit{(ii)} support for arbitrary output resolutions, \textit{(iii)} compatibility with any vision encoder, and \textit{(iv)} minimal computational overhead at inference time.

To enable upsampling to arbitrary target resolutions, we formulate our approach as a global interpolation mechanism using a cross-attention block. The success of this attention-based method depends critically on achieving strong semantic alignment between the queries and keys. In JAFAR, we construct these representations asymmetrically (see \cref{fig:jafar}): the queries retain high-resolution, low-level details such as color and texture, while the keys are hybrid features that combine high-level semantics with spatial cues. We find that enriching the keys with low-level information significantly improves query-key alignment and enhances generalization to unseen output resolutions.

Additionally, we propose a simple training objective similar to~\cite{lift}, but without being constrained to a fixed upsampling factor. Notably, we find that training on low upsampling factors at low resolutions (e.g., $8 \times 8 \rightarrow 32 \times 32$) is sufficient to generalize effectively to much larger scales (e.g., $32 \times 32 \rightarrow 448 \times 448$) while keeping memory requirements low during training, unlike training directly at higher resolutions and factors. Our contributions can be summarized as follows:

\begin{itemize}
    \item We introduce JAFAR, a novel lightweight attention-based feature upsampler that naturally supports upsampling to arbitrary resolutions. It explicitly promotes spatial alignment between high-resolution queries extracted from low-level image features and semantically enriched low-resolution keys.

    \item We enforce this alignment by computing both queries and keys from the same input features, and injecting semantic information from the encoder’s deep features via spatial feature modulation. This design enables precise fusion of spatial detail and semantic context without reliance on external supervision.

    \item We propose a highly efficient, task-agnostic training objective that requires no high-resolution supervision signal. Remarkably, we show that training at low resolutions and low upsampling ratios generalizes robustly to significantly higher output scales.

    \item We demonstrate that the combination of our architecture and training objective yields substantial performance gains across a variety of downstream tasks. When used as a drop-in module, JAFAR consistently outperforms existing upsampling methods by a wide margin.

\end{itemize}

\section{Related Work}
\label{section:related_work}

\paragraph{Feature Upsampling}
Feature upsampling aims to increase the spatial resolution of intermediate feature maps within deep networks—analogous to image upsampling, but performed in a latent space. This process is essential for dense prediction tasks such as segmentation and depth estimation, where fine spatial detail is critical. Traditional interpolation techniques, such as bilinear, spline, or Lanczos \cite{maeland1988comparison, lanczos, cardinalspline, cubic}, provide simple and efficient baselines but do not adapt to the underlying content. Recent neural methods improve on static approaches by learning to reconstruct high-resolution features from data. These methods fall into two categories: task-dependent, trained with downstream labels supervision, and task-agnostic, trained independently of the end task. For example, CARAFE \cite{carafe} and DySample \cite{dysample} predict content-aware kernels or dynamic sampling positions. SAPA \cite{sapa} and ReSFU \cite{resfu} exploit a similarity based approach to refine spatial semantics. However, task-specific reliance on labels limits generalization. Recent task-agnostic methods like LiFT \cite{lift} and FeatUp \cite{featup} remove this dependency. LiFT introduces a CNN module trained with a simple fixed scale training objective, while FeatUp relies on a complex multi-loss objective which makes training difficult to tune in practice. Moreover, it requires training both an upsampler and a downsampler, adding unnecessary computational overhead. Notably, its best performance is achieved through per-image optimization, further limiting its practicality. In contrast, JAFAR provides a scalable task-agnostic framework that generalizes across resolutions without complex pipelines or per-image optimization, showing strong performance even when trained on small upsampling factors at low resolution.

\paragraph{Architectural Design for Upsampling Modules}
Upsampling modules architectures vary from fixed-scale decoders to continuous resolution predictors. LiFT \cite{lift} relies on a lightweight CNN module trained to upsample by a fixed factor, making further scaling dependent on iterative use which leads to performance degradation or additional interpolation steps. FeatUp \cite{featup} introduces two architectural variants: a fast Joint Bilateral Upsampler (JBU) and a more accurate implicit network allowing continuous querying. While the implicit model yields superior results, it suffers from significant inference latency due to per-image optimization. JBU, on the other hand, trades expressivity for scalability, stacking multiple $\times2$ stages to achieve higher upsampling ratios. Attention-based designs, as in SAPA \cite{sapa} and ReSFU \cite{resfu}, offer increased flexibility by modeling affinities between features across scales. These methods exploit spatial similarities to reconstruct high-resolution maps. JAFAR innovates by unifying low- and high-resolution streams: it aligns high-resolution queries and low-resolution keys using shared low-level features while enriching the representation with additional semantic cues. This design maintains spatial alignment and expressivity even at large upsampling ratios, offering a robust and scalable architecture for feature reconstruction.

\paragraph{Semantic Guidance and Feature Modulation}
Feature modulation techniques modulate features using conditioning information, thereby enabling spatially or semantically guided transformations. Early forms such as Conditional BatchNorm \cite{condbatchnorm}, AdaIN \cite{adain}, and FiLM \cite{film} apply learned scale ($\gamma$) and shift ($\beta$) parameters per channel, derived from global conditioning signals. These methods are effective for tasks involving global transformations like style transfer or classification. However, their spatial invariance limits expressiveness in tasks requiring spatial sensitivity. SPADE \cite{spade} and SFT \cite{sft} address this limitation by computing $\gamma$ and $\beta$ as full-resolution maps conditioned on dense inputs like segmentation masks. This spatial modulation enhances expressivity by enabling unique adjustments at each feature location. It can be interpreted as a parameterized, learned recombination of feature channels, analogous to a $1\times1$ convolution but extended with spatially varying weights.  In JAFAR, feature modulation is used not only to shift feature distributions but also to inject semantics directly into the upsampling pipeline. This provides richer linear combinations of features, improving generalization and spatial fidelity without the need for per-image optimization \cite{featup}.

\section{JAFAR}
\label{section:jafar} 

JAFAR is a feature upsampler that uses the input image as high-resolution guidance to reconstruct dense feature maps. To support upsampling to arbitrary target resolutions, we formulate the method as a global interpolation mechanism based on cross-attention. The effectiveness of this attention-based approach hinges on achieving strong semantic alignment between the queries $Q$ and the keys $K$. In JAFAR, we construct the query and key representations asymmetrically. The queries retain high-resolution, low-level details such as color and texture, while the keys are designed as hybrid representations that combine high-level semantics with low-level spatial cues. We find that enriching the keys with low-level information significantly improves query-key alignment and enhances generalization to unseen output resolutions.

\subsection{Architecture}

\begin{figure}[!ht]
\makebox[1.\textwidth]{
  \centering
  \includegraphics[scale=0.62]{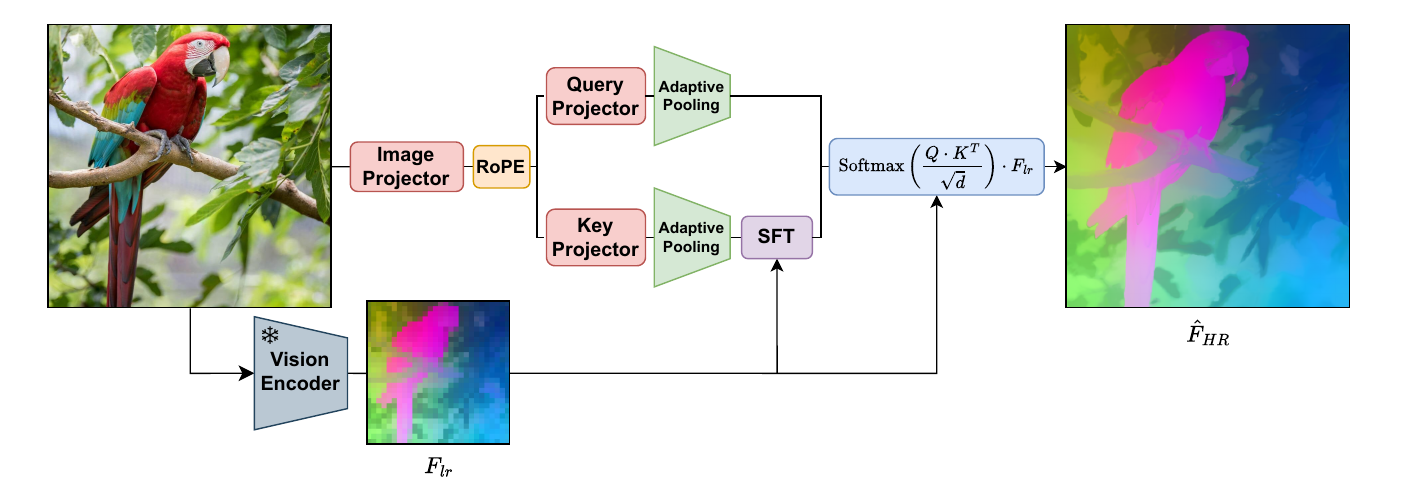}}
  \caption{\textbf{Overview of JAFAR.} To construct the upsampling kernel, queries and keys are derived from a shared image representation. Queries are downsampled to match the target output resolution, while keys are downsampled to align with the spatial resolution of the vision encoder’s features. Keys are then semantically enriched via SFT modulation to promote semantic alignment between queries and keys. The resulting kernel is then used to interpolate features from the foundation vision encoder.}
  \label{fig:jafar}
\end{figure}

The overall flow of our architecture is illustrated in \cref{fig:jafar}. JAFAR takes as input a high-resolution image \( I \in \mathbb{R}^{3 \times H \times W} \) and a low-resolution feature map \( F_{lr} = f(I) \in \mathbb{R}^{C \times h_k \times w_k} \), extracted from a frozen vision encoder \( f \). The image \( I \) is first projected into a higher-dimensional space and processed by a lightweight encoder $E_\theta$ to obtain an intermediate representation $I_E = E_\theta(I) \in \mathbb{R}^{d \times H \times W}$, further enriched with RoPE positional embeddings \cite{rope}. 

Query features $Q \in \mathbb{R}^{d \times h_q \times w_q}$ are derived by passing the image representation $I_E$ through a small query encoder, producing $I_Q$, followed by adaptive average pooling to reach the target resolution $(h_q \times w_q)$. Key features $K \in \mathbb{R}^{d \times h_k \times w_k}$ are similarly obtained by encoding $I_E$ to $I_K$ and downsampling it to match the spatial resolution of the semantic features $F_{lr}$. These semantic features provide modulation parameters that inject high-level information into the keys. A cross-attention mechanism then enables the queries $Q$ to attend to the keys $K$ by computing an attention map:
\begin{equation}
    A = \text{Softmax}\left(\frac{Q \cdot K^\top}{\sqrt{d}}\right),
\end{equation}
which is then used to interpolate the low-resolution feature map $F_{lr}$ and produce the upsampled output features $\hat{F}_{HR} = A \cdot F_{lr} \in \mathbb{R}^{C \times h_q \times w_q}$. The resulting representation preserves fine-grained spatial details while remaining semantically consistent with the input image. We provide a detailed description of each of the main components of the architecture below.

\paragraph{Query Branch} Directly aligning high-resolution, low-level queries with high-level semantic keys often results in weak or noisy attention, as the disparity in abstraction levels limits meaningful interactions. To overcome this challenge, we apply adaptive average pooling to downsample the intermediate representation $I_Q$ and generate the query features $Q$. This operation, performed exclusively during training, reduces the spatial resolution of the queries while aggregating local context into region-level descriptors. As a result, the downsampled queries are more semantically aligned with the keys, less susceptible to pixel-level noise, and computationally more efficient due to the reduced number of tokens. These effects collectively make query downsampling an effective strategy for bridging the gap between fine-grained visual details and abstract semantic representations, promoting more stable and scalable cross-scale attention. Importantly, because downsampling is only applied during training, the model maintains its capacity to generate high-resolution outputs during inference.

\paragraph{Key Branch} 

Relying exclusively on low-resolution features from the vision encoder to construct keys leads to poor generalization and noticeable artifacts, primarily due to an abstraction gap between these coarse features and the fine-grained queries. As demonstrated in \cref{section:experiments}, this mismatch results in inconsistent alignment across resolutions. To address this issue, we construct hybrid key representations that retain structural alignment with the queries while incorporating the semantic richness of the vision encoder. This is achieved by encoding the intermediate representation $I_E$ to produce $I_K$, which is then downsampled to match the spatial resolution of the encoder’s feature map to produce preliminary keys $\tilde{K}$. These are further modulated using the vision encoder feature map $F_{lr} \in \mathbb{R}^{C \times h_k \times w_k}$ through a spatial semantic feature modulation inspired by \cite{spade, sft}:
\begin{equation}
    K = \gamma_F \odot \tilde{K} + \beta_F,
\end{equation}
where $\gamma_F, \beta_F \in \mathbb{R}^{d \times h_k \times w_k}$ are spatially varying parameters obtained via linear projections from $F_{lr}$. This adaptive, feature-wise modulation enriches the keys with localized semantic context, enhancing both spatial and semantic alignment and supporting more faithful and generalizable upsampling across resolutions.

\paragraph{Similarity Based Upsampling} To perform upsampling, we use a simplified attention mechanism where attention weights are computed via a scaled dot product between queries and semantically modulated keys. Crucially, both queries and keys have been enriched with relative positional embeddings using RoPE ~\cite{rope}, which introduces an inductive bias that captures spatial relationships between queries and keys. This positional encoding allows us to entirely bypass the arbitrary selection of neighboring keys for each query, a common heuristic in prior similarity-based methods such as~\cite{sapa, resfu}.  Without this positional grounding, the attention mechanism lacks spatial awareness and generalizes poorly to unseen resolutions. In practice, we use multiple attention heads to increase expressivity and average the resulting attention weights across heads after applying softmax. The resulting attention map $A$ is then used to interpolate the low-resolution encoder features $F_{lr}$ via a simple matrix product: $\hat{F}_{HR} = A \cdot F_{lr}$. By avoiding a learned value projection, we preserve the original feature content and enable a resolution-agnostic design that generalizes reliably across scales.

\subsection{Training Pipeline}
Learning to upsample high-resolution features without access to ground-truth supervision poses a natural challenge: how can a model learn to produce sharp high-resolution features (e.g., $448 \times 448$) when only low-resolution features are available (e.g., $32 \times 32$)? Thanks to JAFAR’s architectural design, the model can be trained with a simple objective at a low target resolution without requiring supervision at the original image size, yet it still generalizes effectively to much higher upsampling ratios during inference.

\paragraph{Training with Multi-Resolution Views}

 To enable this, we introduce a fully annotation-free training scheme that relies only on multi-resolution views of the same image, easily obtained through standard downsampling. Given a high-resolution image $I_{HR} \in \mathbb{R}^{3 \times H \times W}$, we generate a downsampled version $I_{LR} \in \mathbb{R}^{3 \times \lfloor\frac{H}{\delta}\rfloor \times \lfloor\frac{W}{\delta}\rfloor}$ using a randomly sampled factor $\delta \in [2, 4]$. Both images are passed through the frozen vision encoder $f$, producing two feature maps: $F_{hr} = f(I_{HR}) \in \mathbb{R}^{C \times h \times w}$ and $F_{lr} = f(I_{LR}) \in \mathbb{R}^{C \times \lfloor\frac{h}{\delta}\rfloor  \times \lfloor\frac{w}{\delta}\rfloor }$, respectively. JAFAR then takes $I_{HR}$ and $F_{lr}$ as input to predict an upsampled feature map $\hat{F}_{hr}$. The predicted output is aligned with the target $F_{hr}$ using a simple alignment loss, which combines cosine similarity and L2 distance \cite{dvt}: 
\begin{equation}
    \mathcal{L}(\hat{F}_{hr}, F_{hr}) = 1 - \text{cos}(\hat{F}_{hr}, F_{hr}) + ||\hat{F}_{hr} - F_{hr}||_2.
\end{equation}
Notably, during training, JAFAR is only exposed to moderate upsampling factors (up to $4\times$), yet it generalizes remarkably well to much higher resolutions at test time—without access to any ground-truth high-resolution features.

\paragraph{How is it different from LiFT?} While our training objective is similar to that of LiFT, our approach demonstrates significantly greater capability, as shown in \cref{table:probing_dinov2_supervised,table:cam_table}. LiFT relies on a CNN-based architecture and is trained for fixed $2\times$ upsampling at two predefined resolutions. As a result, it struggles to extrapolate beyond that setting without additional heuristics such as iterative upsampling or bilinear fallback. In contrast, JAFAR maintains a resolution-agnostic design which generalizes to much higher upsampling factors using this similar simple training setup.

\section{Experiments}
\label{section:experiments}

\subsection{Experimental Setup}

In our experiments, we train JAFAR on a single NVIDIA A100 on ImageNet training set for 100K steps using AdamW optimizer \cite{adamw}, with a learning rate of $2\text{e}{-4}$ and a batch size of 4. The input images fed into the foundation vision encoder are resized to $448 \times 448$, producing high-resolution target feature maps $F_{hr}$ of size $32 \times 32$ or $28 \times 28$, depending on the encoder’s patch size (14 or 16). For improved training efficiency, the guidance image input to JAFAR is downsampled to $224 \times 224$. 

\subsection{Qualitative Comparisons}

\begin{figure}[!ht]
\makebox[1.\textwidth]{
  \centering
  \includegraphics[scale=0.58]{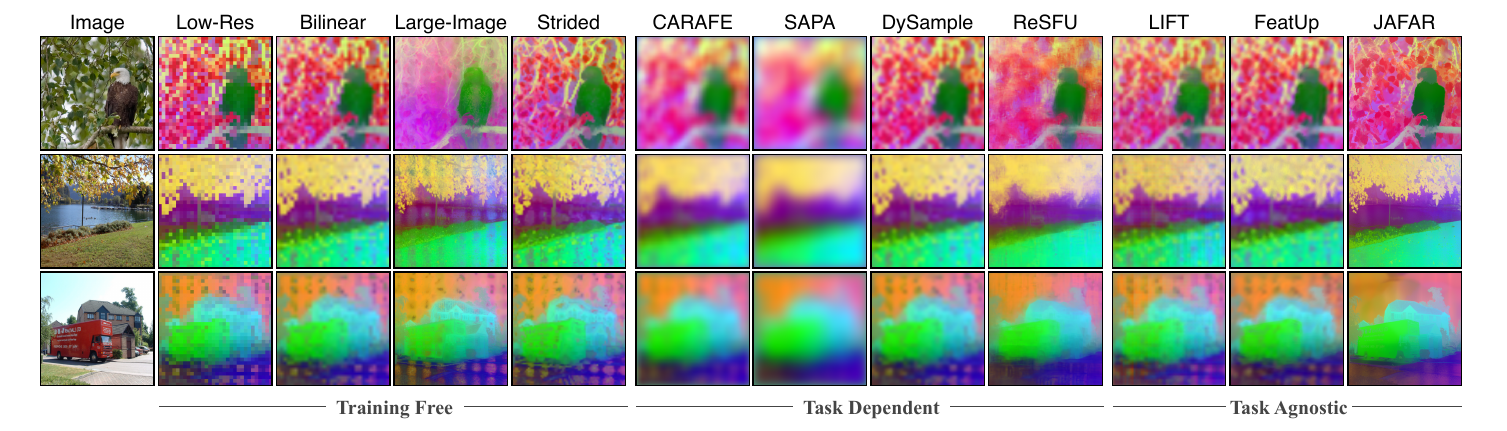}}
  \caption{\textbf{PCA Feature Visualization.}
DINOv2 ViT-S/14 features at $32 \times 32$ resolution from the ImageNet validation set are upsampled to $448 \times 448$. Baseline methods—whether training-free, task-dependent, or task-agnostic—introduce varying levels of blurriness and artifacts. Besides being task-agnostic, JAFAR produces sharp, content-aware feature maps with fewer artifacts.}

  \label{fig:features_comparison}
\end{figure}

To qualitatively evaluate the upsampled feature maps produced by various baselines, we project all features onto a shared 3-dimensional PCA basis, mapping them into a common RGB space. As shown in \cref{fig:features_comparison,fig:features_comparison_supp}, the low-resolution features—due to the spatial compression imposed by the vision encoder’s patch size—reveal large, blocky regions that capture semantic content but fail to preserve fine image geometry, object boundaries, or shape details. Bilinear upsampling, which interpolates features without considering image content, yields blurry output feature maps that preserve positional embeddings artifacts without adding meaningful detail. While methods like Large-Image and Strided preserve sharpness, their outputs are noisier and less coherent than JAFAR’s. Furthermore, they are more computationally demanding, as they require the vision encoder to process a larger number of patches (see \cref{tab:params_and_speed}). JAFAR shows a clear qualitative advantage over all baselines, consistently producing sharp features that accurately capture image structure. It is also the only task-agnostic method that effectively suppresses artifacts from positional embeddings in the low-resolution features.

\subsection{Transfer on Downstream Tasks}

Since upsampled features are expected to provide a richer signal for downstream tasks, we evaluate their effectiveness on two benchmarks: linear-probing semantic segmentation and depth estimation, using DINOv2 ViT-S/14 as the foundation vision encoder. For the Large-Image and Strided baselines, upsampling is performed during the encoder’s forward pass and followed by bilinear interpolation to reach the target output resolution. For task-agnostic upsamplers such as LiFT, FeatUp, and JAFAR, we pre-train the upsampling module on the corresponding backbone, then freeze it and apply it after feature extraction. The linear probe is trained independently of the upsampler. For task-dependent methods, including CARAFE, SAPA, ReSFu, and DySample, we jointly train both the upsampler and the linear probe on each dataset and task. All experiments (except Large-Image) use input images of resolution $448 \times 448$, with target labels at the same resolution.

\subsubsection{Semantic Segmentation}
For semantic segmentation, we train a linear projection head to predict coarse class labels using a cross-entropy loss across several benchmark datasets: COCO-Stuff \cite{lin2014microsoft} (27 classes), ADE20K \cite{ade20k} (150 classes), Pascal VOC \cite{pascalvoc} (21 classes including background), and Cityscapes \cite{cityscapes} (27 classes). The linear layer is trained for 5 epochs on COCO-Stuff and 20 epochs on the remaining datasets, using a batch size of 4. Performance is evaluated on the respective validation sets using mean Intersection-over-Union (mIoU) and pixel-wise accuracy.

\begin{table}[!h]
\vspace{-5pt}
\centering
\caption{\textbf{Linear Probing on Downstream Tasks.} JAFAR consistently outperforms other baselines across all segmentation benchmarks while reaching competitive depth metrics without being optimized on a specific downstream task.}
\scalebox{0.73}{
\makebox[\textwidth]{
\begin{tabular}{lcccccccccc}
    \toprule
    \multirow{2}{*}{\textbf{DINOv2-ViT-S/14}} & \multicolumn{8}{c}{\textbf{Semantic Segmentation}} & \multicolumn{2}{c}{\textbf{Depth Estimation}} \\
    \cmidrule(rl){2-9} \cmidrule(rl){10-11}
    & \multicolumn{2}{c}{\textbf{COCO}} & \multicolumn{2}{c}{\textbf{VOC}} & \multicolumn{2}{c}{\textbf{ADE20K}} & \multicolumn{2}{c}{\textbf{Cityscapes}} & \multicolumn{2}{c}{\textbf{COCO}} \\
    \cmidrule(rl){2-3} \cmidrule(rl){4-5} \cmidrule(rl){6-7} \cmidrule(rl){8-9} \cmidrule(rl){10-11}
    & mIoU ($\uparrow$) & Acc ($\uparrow$) & mIoU ($\uparrow$) & Acc ($\uparrow$) & mIoU ($\uparrow$) & Acc ($\uparrow$) & mIoU ($\uparrow$) & Acc ($\uparrow$) & $\delta_1$ ($\uparrow$) & RMSE ($\downarrow$) \\
    \cmidrule(rl){1-1} \cmidrule(rl){2-11}
    \textit{\textbf{Training-free}} \\
    
    \quad Nearest & 56.17 & 76.97 & 76.41 & 93.80 & 37.27 & 71.91 & 54.05 & 90.36 & 58.08 & 0.70 \\
    
    \quad Bilinear & 59.03 & 79.07 & 80.70 & 95.17 & \underline{39.23} & 73.69 & 59.37 & 92.47 & 59.92 & 0.66 \\
    
    \quad Large Image (x8) & -- & -- & 56.94 & 88.60 & 26.42 & 66.39 & 47.72 & 92.49 & -- & -- \\
    
    \quad Strided & 55.93 & 77.40 & 75.88 & 93.94 & 36.15 & 72.08 & 59.26 & 92.57 & 56.98 & 0.70 \\
    
    \cmidrule(rl){1-1} \cmidrule(rl){2-11}

    \textit{\textbf{Task-Dependent}} \\

    \quad CARAFE \cite{carafe}   & 59.73 & 79.65 & 80.26 & 95.14 & 38.30 & 73.42 & 56.05 & 91.83 & 61.42 & 0.64 \\
    
    \quad SAPA \cite{sapa}       & 57.77 & 78.28 & 77.02 & 94.07 & 35.87 & 71.85 & 50.12 & 90.02 & 60.34 & 0.67 \\
    
    \quad DySample \cite{dysample} & 59.50 & 79.42 & \underline{81.62} & \underline{95.48} & 38.99 & 73.62 & \underline{59.71} & \underline{92.69} & 61.25 & 0.64 \\
    
    \quad ReSFU \cite{resfu}     & 60.08 & 79.84 & 80.30 & 95.05 & 38.91 & \underline{73.93} & 55.53 & 91.62 & \textbf{66.14} & \textbf{0.56} \\
    \cmidrule(rl){1-1} \cmidrule(rl){2-11}

    \textit{\textbf{Task-Agnostic}} \\
    
    \quad FeatUp \cite{featup} & \underline{60.10} & \underline{79.95} & 81.08 & 95.32 & 38.82 & 73.74 & 56.06 & 91.86 & 61.69 & 0.64 \\
    
    \quad LIFT \cite{lift} & 58.18 & 78.95 & 78.06 & 94.62 & 38.73 & 73.69 & 58.75 & 92.60 & 57.04 & 0.70 \\
    \cmidrule(rl){1-1} \cmidrule(rl){2-11}
    
    \quad JAFAR & \textbf{60.78} & \textbf{80.47} & \textbf{84.44} & \textbf{96.28} & \textbf{40.49} & \textbf{74.92} & \textbf{61.47} & \textbf{93.42} & \underline{62.18} & \underline{0.62} \\

    \bottomrule
\end{tabular}}}
\label{table:probing_dinov2_supervised}
\end{table}

As shown in \cref{table:probing_dinov2_supervised}, JAFAR consistently achieves the highest performance across all four semantic segmentation benchmarks, in both mIoU and accuracy. On average, JAFAR delivers a $+1.63$ mIoU improvement over the next-best method across all datasets. Compared to FeatUp, JAFAR achieves an average gain of $+2.78$ mIoU corresponding to a $+4.8\%$ gain, with a peak improvement of $+5.41$ mIoU ($+9.7\%$) on Cityscapes. \cref{fig:features_comparison_tasks} shows linear probe segmentation result.

\subsubsection{Depth Estimation}

\begin{figure}[!ht]
\makebox[1.\textwidth]{
  \centering
  \includegraphics[scale=0.56]{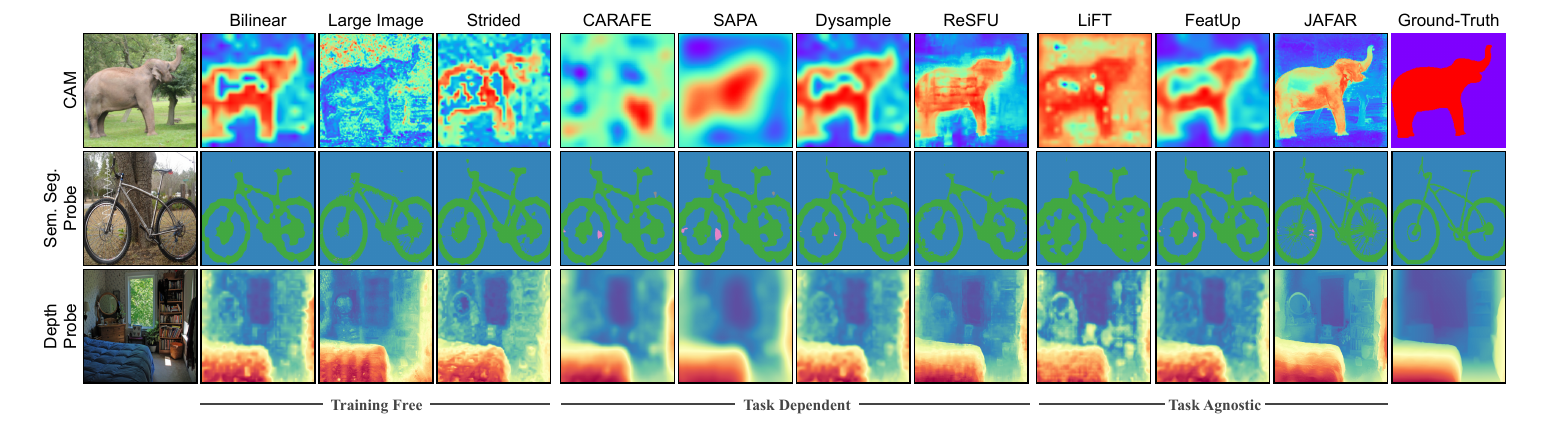}}
  \caption{\textbf{Visual Comparison of Upsampler Outputs in Downstream Tasks.} JAFAR-upsampled features produce sharper outputs that align more accurately with object boundaries across various downstream tasks respectively class activations maps, semantic segmentation and depth estimation.}
  \label{fig:features_comparison_tasks}
\end{figure}

For depth estimation, we follow the approach in \cite{featup} and train on pseudo-labels generated by the state-of-the-art Depth Anything V2 network \cite{depth_anythingv2}. We report two standard metrics from the monocular depth estimation literature: root mean square error (RMSE) and $\delta_1 < 1.25$. The $\delta_1$ metric measures the percentage of pixels where the predicted depth $y$ is within 25\% of the ground-truth $y^*$, formally defined as $\delta_1 = max\left(\frac{y}{y^*}, \frac{y^*}{y}\right) < 1.25$. We train the linear probe for 5 epochs on the COCO training set, using a batch size of 4. Although JAFAR was not trained on this specific task, we observe that it reaches competitive scores, ranking second among the baselines. Notably, JAFAR outperforms both FeatUp and LiFT while also surpassing all task-dependent methods but ReSFU. \cref{fig:features_comparison_tasks} shows linear probe depth estimation result. 

\subsubsection{Class Activation Maps Faithfulness}

Following the approach in \cite{featup}, our method can be seamlessly integrated into explainability tools such as Class Activation Maps (CAMs). Despite recent advances, CAMs are still fundamentally limited by the low-resolution feature maps produced by standard vision encoders, which hinders their ability to localize fine-grained details. By upsampling the features, our method yields sharper and more informative explanations. To assess the quality of the resulting CAMs, we adopt standard evaluation metrics from the literature: Average Drop (A.D), Average Increase (A.I), Average Gain (A.G), Coherency (Coh.), and Complexity (Cplx.). 

\begin{wraptable}[16]{r}{0.6\textwidth}
\centering
\caption{\textbf{Grad-CAM Evaluation.} Integrating JAFAR into Grad-CAM analysis yields significantly more faithful explanations compared to baseline methods. Top three methods are highlighted as \textcolor{gold}{first}, \textcolor{silver}{second}, \textcolor{bronze}{third} according to \textbf{ADCC}.}
\small
\scalebox{0.7}{
\makebox[0.7\textwidth]{
\begin{tabular}{lccccc|l}
\toprule
& \textbf{A.D} ($\downarrow$) & \textbf{A.I} ($\uparrow$) & \textbf{A.G} ($\uparrow$) & \textbf{Coh.} ($\uparrow$) & \textbf{Cplx.} ($\downarrow$) & \textbf{ADCC} ($\uparrow$)\\
\cmidrule(rl){1-1} \cmidrule(rl){2-7}
\textit{\textbf{Training-free}} \\
\quad Bilinear & 19.0 & 18.5 & 3.4 & 88.8 & 59.9 & 61.7 \scriptsize\textcolor{red}{(-11.6)}  \\
\quad Large Image (x8) & 48.8 & 12.8 & 2.5 & 67.9 & 38.0 & 59.5 \scriptsize\textcolor{red}{(-13.8)} \\
\rowcolor{silver!20} \quad Strided & 19.8 & 15.1 & 3.5 & 85.9 & 54.2 & 65.3 \scriptsize\textcolor{red}{(-8.0)} \\
\cmidrule(lr){1-1}\cmidrule(lr){2-7}
\textit{\textbf{Task-Dependent}} \\
\quad CARAFE \cite{carafe} & 49.9 & 4.8 & 1.0 & 66.9 & 34.9 & 59.7 \scriptsize\textcolor{red}{(-13.6)} \\
\quad SAPA \cite{sapa} & 8.5 & 32.7 & 4.1 & 96.5 & 69.6 & 55.4 \scriptsize\textcolor{red}{(-17.9)} \\
\quad Dysample \cite{dysample} & 17.8 & 20.0 & 3.8 & 90.2 & 60.7 & 61.6 \scriptsize\textcolor{red}{(-11.7)} \\
\quad ReSFU \cite{resfu} & 14.5 & 24.0 & 3.7 & 92.1 & 64.2 & 59.4 \scriptsize\textcolor{red}{(-13.9)} \\
\cmidrule(lr){1-1}\cmidrule(lr){2-7}
\textit{\textbf{Task-Agnostic}} \\
\rowcolor{bronze!20}  \quad FeatUp \cite{featup} & 15.3 & 24.0 & 4.3 & 91.6 & 58.2 & 64.3 \scriptsize\textcolor{red}{(-9.0)} \\
\quad LiFT \cite{lift} & 66.9 & 8.7 & 2.3 & 65.2 & 9.4 & 53.0 \scriptsize\textcolor{red}{(-20.3)} \\
\rowcolor{gold!20} \quad JAFAR & 17.4 & 30.9 & 6.5 & 91.4 & 44.1 & 73.3\\
\bottomrule
\end{tabular}}}
\label{table:cam_table}
\end{wraptable}

In particular, A.D, A.I, and A.G measure how sensitive the classifier’s output is to the most salient regions of the input—an effective CAM should highlight areas that, when masked, lead to a notable change in classification confidence. Since each of these metrics captures only a single aspect of CAM quality, we also report the ADCC score—an aggregate metric proposed in \cite{adcc} that provides a more holistic evaluation. Additional details are provided in \cref{section:cam_evaluation}. As illustrated qualitatively in \cref{fig:features_comparison_tasks}, JAFAR generates sharper and more semantically accurate CAMs compared to all baselines. While training-free methods don't help to recover important regions, task-dependent approaches typically produce blurrier and less precise maps. Quantitative results in \cref{table:cam_table} further support this, with JAFAR achieving the highest score on the aggregate ADCC metric—outperforming the second-best method by 8 points, a relative improvement of 12.5\%.

\subsubsection{Zero-Shot Open-Vocabulary Segmentation}

We further evaluate our method on a zero-shot open-vocabulary segmentation task, following the setup from \cite{maskclip}, where class labels from the dataset serve as textual inputs and predictions are made by selecting the class with the highest similarity score (argmax). Using a CLIP-ViT-B/16 backbone, this approach is entirely training-free, as it does not require a learned probing head. Results show that JAFAR significantly outperforms all baselines, with particularly strong improvements on Pascal VOC. Despite the increased difficulty of ADE20K, which includes 150 classes, our method still achieves the highest performance in both mIoU and accuracy. We report only FeatUp among the task-agnostic baselines, as it is the second-best performing method.

\begin{table}[ht]
\centering
\captionsetup{width=0.82\textwidth}
\caption{\textbf{Zero-Shot Open-Vocabulary Evaluation.} Using MaskCLIP \cite{maskclip} for zero-shot open-vocabulary segmentation, JAFAR consistently improves performance, indicating strong alignment with the original features.}
\label{tab:maskclip_upsampling_all_column}
\scalebox{0.9}{
\makebox[\textwidth]{
\begin{tabular}{llcccccc}
\toprule
 & 
\multirow{2}{*}{} & 
\multicolumn{2}{c}{\textbf{VOC}} & 
\multicolumn{2}{c}{\textbf{ADE20K}} & 
\multicolumn{2}{c}{\textbf{Cityscapes}} \\
\cmidrule(lr){3-4} \cmidrule(lr){5-6} \cmidrule(lr){7-8}
\textbf{Upsampling} & & mIoU ($\uparrow$) & Acc ($\uparrow$) & mIoU ($\uparrow$) & Acc ($\uparrow$) & mIoU ($\uparrow$) & Acc ($\uparrow$) \\
\midrule
Nearest              & & 24.13 & 30.80 & 9.33 & 24.65 & 19.66 & 50.27 \\
Bilinear             & & 27.87 & 35.27 & 11.03 & 27.78 & 21.56 & 53.21 \\
Large Image ($\times2$) &  & 23.24 & 32.16 & 8.08 & 24.94 & 21.91 & 52.22 \\
FeatUp \cite{featup} & & \underline{32.27} & \underline{39.78} &  \underline{13.03} & \underline{33.28} & \underline{24.76} & \underline{60.11} \\
JAFAR                & & \textbf{35.70} & \textbf{44.93} & \textbf{13.61} & \textbf{33.28} & \textbf{25.26} & \textbf{61.73} \\
\bottomrule
\end{tabular}}}
\end{table}

\subsubsection{Bird’s-Eye View Segmentation}

\begin{wraptable}[8]{r}{0.5\textwidth}
\vspace{-12pt}
\centering
\captionsetup{width=0.49\textwidth}
\caption{\textbf{BeV Vehicle Segmentation.} JAFAR consistently improves vehicle-IoU in complex BeV architectures, outperforming all other baselines.}
\label{tab:bev_segmentation_upsampling_column}
\small
\scalebox{0.7}{
\makebox[0.7\textwidth]{
\begin{tabular}{lccc}
\toprule
\textbf{Upsampling} & \textbf{SimpleBeV} \cite{harley2023simple} & \textbf{PointBeV} \cite{pointbev} & \textbf{BeVFormer} \cite{bevformer} \\
\midrule
Low-Res & 31.75 & 34.89 & 33.72 \\
Bilinear  & 33.67 & \underline{36.01} & \underline{34.18} \\
FeatUp    & \underline{33.95} &  35.38 & 34.01 \\
JAFAR     & \textbf{36.59} & \textbf{37.20} & \textbf{36.54} \\
\bottomrule
\end{tabular}}}
\end{wraptable}

Finally, we studied the impact of our upsampler in a complex training pipeline. The task, evaluated on nuScenes \cite{nuscenes}, takes several images taken from cameras as input and consists on outputting the bird's-eye view (BeV) segmentation map. In our setup, we used a frozen DINOv2 \cite{dinov2} backbone and trained the rest of the architecture, namely, the upsampler, the BeV encoder, and the segmentation head. This task is particularly challenging, as the model must learn to map features from the image plane to the BeV plane. To ensure a fair comparison, we also trained the architecture without an upsampler, using lower-resolution input images $(224 \times 476)$. We adopted the optimization hyperparameters from PointBeV \cite{pointbev}, adjusting the batch size to 1 and training for 100 epochs. Our results show that using an upsampler consistently improves predictions, regardless of the architecture employed—SimpleBev \cite{harley2023simple}, PointBeV \cite{pointbev}, or BevFormer \cite{bevformer}. Notably, performance improves significantly when using JAFAR as the upsampler, with mIoU gains up to $+5$ points.

\subsection{Ablations}

To evaluate the benefit of deriving both queries and keys from a shared image encoding, we compare in \cref{table:ablation_attention_merged} several strategies to obtain keys. In the Linear Projection baseline, keys are obtained by applying a linear layer to the vision encoder's low-resolution features $F_{lr}$, to match JAFAR's embedding dimension. In the Concatenation baseline, semantics is injected via a direct concatenation of $F_{lr}$ and preliminary keys $\tilde{K}$.
\begin{table}[!h]
\definecolor{lightpastelblue}{rgb}{0.75, 0.85, 0.95}
\vspace{-10pt}
\centering
\captionsetup{width=0.92\textwidth}
\caption{\textbf{Attention mechanism ablations} with respect to key strategy and number of attention heads. Best scores per dataset are in bold and selected choices are highlighted in blue.}
\scalebox{0.82}{
\makebox[\textwidth]{
\begin{tabular}{llllllll}
    \toprule
    \multirow{2}{*}{\textbf{Ablation Type / Setting}} & \multicolumn{6}{c}{\textbf{Semantic Segmentation}}\\
    \cmidrule(rl){2-7}
    & \multicolumn{2}{c}{\textbf{VOC}} & \multicolumn{2}{c}{\textbf{ADE20K}} & \multicolumn{2}{c}{\textbf{Cityscapes}}\\
    \cmidrule(rl){2-3} \cmidrule(rl){4-5} \cmidrule(rl){6-7}
    & \multicolumn{1}{c}{mIoU ($\uparrow$)}  & \multicolumn{1}{c}{Acc ($\uparrow$)}  & \multicolumn{1}{c}{mIoU ($\uparrow$)}  & \multicolumn{1}{c}{Acc ($\uparrow$)} & \multicolumn{1}{c}{mIoU ($\uparrow$)} & \multicolumn{1}{c}{Acc ($\uparrow$)} \\
    \cmidrule(rl){1-1} \cmidrule(rl){2-7}
    \textit{\textbf{Keys Strategy}} \\
    \quad Linear Projection & 80.02 \scriptsize \textcolor{red}{(-4.42)}& 94.87 \scriptsize\textcolor{red}{(-1.41)} & 37.87 \scriptsize \textcolor{red}{(-2.62)} & 73.22 \scriptsize \textcolor{red}{(-1.70)} & 52.45 \scriptsize \textcolor{red}{(-9.02)}& 90.80 \scriptsize \textcolor{red}{(-2.62)}\\
    \quad Concatenation & 83.13 \scriptsize \textcolor{red}{(-1.27)}& 95.94 \scriptsize \textcolor{red}{(-0.34)} & 40.06 \scriptsize \textcolor{red}{(-0.43)} & 74.56 \scriptsize\textcolor{red}{(-0.36)}&  58.70 \scriptsize \textcolor{red}{(-2.77)}& 92.65 \scriptsize \textcolor{red}{(-0.77)}\\
    \quad w/o SFT & 83.25 \scriptsize \textcolor{red}{(-1.19)}& 95.93 \scriptsize \textcolor{red}{(-0.35)} & 39.62 \scriptsize \textcolor{red}{(-0.87)} & 74.32 \scriptsize\textcolor{red}{(-0.60)}&  56.53 \scriptsize \textcolor{red}{(-4.94)}&  92.19 \scriptsize \textcolor{red}{(-1.23)}\\
    \rowcolor{lightpastelblue} \quad w/ SFT & \textbf{84.44} & \textbf{96.28} & \textbf{40.49} & \textbf{74.92} & \textbf{61.47} & \textbf{93.42} \\
    \midrule
    \textit{\textbf{Attention Heads}} \\
    \quad $n = 1$ & 84.13 \scriptsize\textcolor{red}{(-0.31)} & 96.21 \scriptsize\textcolor{red}{(-0.07)} & 40.15 \scriptsize\textcolor{red}{(-0.34)} & 74.79 \scriptsize\textcolor{red}{(-0.13)} & 60.94 \scriptsize\textcolor{red}{(-0.53)} & 93.32 \scriptsize\textcolor{red}{(-0.10)}  \\
    \quad $n = 2$ & 84.27 \scriptsize\textcolor{red}{(-0.17)} & 96.27 \scriptsize\textcolor{red}{(-0.01)} & 40.42 \scriptsize\textcolor{red}{(-0.07)} & \textbf{74.95} \scriptsize\textcolor{green}{(+0.03)} & 61.19 \scriptsize\textcolor{red}{(-0.28)} & 93.42 \scriptsize\textcolor{red}{(-0.00)}\\
    \rowcolor{lightpastelblue} \quad $n = 4$ & \textbf{84.44}  & \textbf{96.28}  & \textbf{40.49} & 74.92 & \textbf{61.47} & \textbf{93.42} \\
    \quad $n = 8$ & {83.82} \scriptsize\textcolor{red}{(-0.62)} & {96.13} \scriptsize\textcolor{red}{(-0.15)} & 40.07 \scriptsize\textcolor{red}{(-0.42)} & 74.20 \scriptsize\textcolor{red}{(-0.72)} & 60.56 \scriptsize\textcolor{red}{(-0.91)} & 93.33 \scriptsize\textcolor{red}{(-0.09)} \\
    \bottomrule
\end{tabular}}}
\vspace{-5pt}
\label{table:ablation_attention_merged}
\end{table}

 In comparison, the Linear projection baseline shows a significant performance drop, and SFT consistently outperforms the concatenation approach. Increasing the number of attention heads up to 4 further enhances performance by producing more robust upsampling kernels through averaged post-softmax scores. Beyond this point, however, the benefits reverse: the per-head dimensionality becomes too low to support effective alignment, while the computational cost increases, ultimately degrading output quality.
\section{Conclusion}
\label{section:conclusion}

We introduce JAFAR, a lightweight, attention-based feature upsampler designed with a simple training objective. It can upscale features from any foundation vision encoder to arbitrary output resolutions, without requiring supervision at the original image size or annotations from downstream tasks. Although task-agnostic, JAFAR outperforms prior state-of-the-art upsamplers across a variety of downstream tasks, despite not being trained specifically for them. This work lays the groundwork for a unified feature upsampler that could enable significantly more efficient architectures for dense vision tasks. 
Currently, the method requires training a separate upsampler for each backbone. Future work will focus on making JAFAR backbone-independent at inference time and on further reducing feature-level artifacts to produce sharper outputs.

\section{Acknowledgment}

This work has been supported by chair VISA DEEP (ANR-20-CHIA-0022) Cluster PostGenAI@Paris (ANR-23-IACL-0007, FRANCE 2030). This work has been supported by PEPR Sharp (ANR-23-PEIA-0008, FRANCE 2030). This work was granted access to the HPC resources of IDRIS under the allocation 2025-AD011014763R1 made by GENCI.

\nocite{*}
\bibliographystyle{unsrt}
\bibliography{bibliography.bib}

@inproceedings{
    featup,
    title={FeatUp: A Model-Agnostic Framework for Features at Any Resolution},
    author={Stephanie Fu and Mark Hamilton and Laura E. Brandt and Axel Feldmann and Zhoutong Zhang and William T. Freeman},
    booktitle={The Twelfth International Conference on Learning Representations},
    year={2024},
    url={https://openreview.net/forum?id=GkJiNn2QDF}
}

@inproceedings{lift,
  title={Lift: A surprisingly simple lightweight feature transform for dense vit descriptors},
  author={Suri, Saksham and Walmer, Matthew and Gupta, Kamal and Shrivastava, Abhinav},
  booktitle={European Conference on Computer Vision},
  pages={110--128},
  year={2024},
  organization={Springer}
}

@article{sapa,
  title={SAPA: Similarity-aware point affiliation for feature upsampling},
  author={Lu, Hao and Liu, Wenze and Ye, Zixuan and Fu, Hongtao and Liu, Yuliang and Cao, Zhiguo},
  journal={Advances in Neural Information Processing Systems},
  volume={35},
  pages={20889--20901},
  year={2022}
}

@inproceedings{carafe,
  title={Carafe: Content-aware reassembly of features},
  author={Wang, Jiaqi and Chen, Kai and Xu, Rui and Liu, Ziwei and Loy, Chen Change and Lin, Dahua},
  booktitle={Proceedings of the IEEE/CVF international conference on computer vision},
  pages={3007--3016},
  year={2019}
}

@article{dysample,
  title={Learning to Upsample by Learning to Sample},
  author={Wenze Liu and Hao Lu and Hongtao Fu and Zhiguo Cao},
  journal={2023 IEEE/CVF International Conference on Computer Vision (ICCV)},
  year={2023},
  pages={6004-6014},
  url={https://api.semanticscholar.org/CorpusID:261276820}
}

@inproceedings{FADE,
  title={FADE: Fusing the Assets of Decoder and Encoder for Task-Agnostic Upsampling},
  author={Lu, Hao and Liu, Wenze and Fu, Hongtao and Cao, Zhiguo},
  booktitle={Proc. European Conference on Computer Vision (ECCV)},
  year={2022}
}

@misc{dinov2,
  title={DINOv2: Learning Robust Visual Features without Supervision},
  author={Oquab, Maxime and Darcet, Timothée and Moutakanni, Theo and Vo, Huy V. and Szafraniec, Marc and Khalidov, Vasil and Fernandez, Pierre and Haziza, Daniel and Massa, Francisco and El-Nouby, Alaaeldin and Howes, Russell and Huang, Po-Yao and Xu, Hu and Sharma, Vasu and Li, Shang-Wen and Galuba, Wojciech and Rabbat, Mike and Assran, Mido and Ballas, Nicolas and Synnaeve, Gabriel and Misra, Ishan and Jegou, Herve and Mairal, Julien and Labatut, Patrick and Joulin, Armand and Bojanowski, Piotr},
  journal={arXiv:2304.07193},
  year={2023}
}

@inproceedings{spade,
  title={Semantic image synthesis with spatially-adaptive normalization},
  author={Park, Taesung and Liu, Ming-Yu and Wang, Ting-Chun and Zhu, Jun-Yan},
  booktitle={Proceedings of the IEEE/CVF conference on computer vision and pattern recognition},
  pages={2337--2346},
  year={2019}
}

@inproceedings{film,
  title={FiLM: Visual Reasoning with a General Conditioning Layer},
  author={Perez, Ethan and Strub, Florian and de Vries, Harm and Dumoulin, Vincent and Courville, Aaron},
  booktitle={AAAI Conference on Artificial Intelligence},
  year={2018}
}

@inproceedings{sft,
  title={Recovering realistic texture in image super-resolution by deep spatial feature transform},
  author={Wang, Xintao and Yu, Ke and Dong, Chao and Loy, Chen Change},
  booktitle={Proceedings of the IEEE conference on computer vision and pattern recognition},
  pages={606--615},
  year={2018}
}

@article{depth_anythingv2,
  title={Depth anything v2},
  author={Yang, Lihe and Kang, Bingyi and Huang, Zilong and Zhao, Zhen and Xu, Xiaogang and Feng, Jiashi and Zhao, Hengshuang},
  journal={Advances in Neural Information Processing Systems},
  volume={37},
  pages={21875--21911},
  year={2024}
}

@Article{pascalvoc,
   author = "Everingham, M. and Eslami, S. M. A. and Van~Gool, L. and Williams, C. K. I. and Winn, J. and Zisserman, A.",
   title = "The Pascal Visual Object Classes Challenge: A Retrospective",
   journal = "International Journal of Computer Vision",
   volume = "111",
   year = "2015",
   number = "1",
   month = jan,
   pages = "98--136",
}

@inproceedings{lin2014microsoft,
  title={Microsoft coco: Common objects in context},
  author={Lin, Tsung-Yi and Maire, Michael and Belongie, Serge and Hays, James and Perona, Pietro and Ramanan, Deva and Doll{\'a}r, Piotr and Zitnick, C Lawrence},
  booktitle={Computer Vision--ECCV 2014: 13th European Conference, Zurich, Switzerland, September 6-12, 2014, Proceedings, Part V 13},
  pages={740--755},
  year={2014},
  organization={Springer}
}

@inproceedings{cityscapes,
  title={The cityscapes dataset for semantic urban scene understanding},
  author={Cordts, Marius and Omran, Mohamed and Ramos, Sebastian and Rehfeld, Timo and Enzweiler, Markus and Benenson, Rodrigo and Franke, Uwe and Roth, Stefan and Schiele, Bernt},
  booktitle={Proceedings of the IEEE conference on computer vision and pattern recognition},
  pages={3213--3223},
  year={2016}
}

@article{ade20k,
  title={Semantic understanding of scenes through the ade20k dataset},
  author={Zhou, Bolei and Zhao, Hang and Puig, Xavier and Xiao, Tete and Fidler, Sanja and Barriuso, Adela and Torralba, Antonio},
  journal={International Journal of Computer Vision},
  volume={127},
  pages={302--321},
  year={2019},
  publisher={Springer}
}

@article{resfu,
  title={A Refreshed Similarity-based Upsampler for Direct High-Ratio Feature Upsampling},
  author={Zhou, Minghao and Wang, Hong and Zheng, Yefeng and Meng, Deyu},
  journal={CoRR},
  year={2024}
}

@inproceedings{
condbatchnorm,
title={A Learned Representation For Artistic Style},
author={Vincent Dumoulin and Jonathon Shlens and Manjunath Kudlur},
booktitle={International Conference on Learning Representations},
year={2017},
url={https://openreview.net/forum?id=BJO-BuT1g}
}

@inproceedings{adain,
  title={Arbitrary style transfer in real-time with adaptive instance normalization},
  author={Huang, Xun and Belongie, Serge},
  booktitle={Proceedings of the IEEE international conference on computer vision},
  pages={1501--1510},
  year={2017}
}

@article{cocostuff,
  title={COCO-Stuff: Thing and Stuff Classes in Context},
  author={Holger Caesar and Jasper R. R. Uijlings and Vittorio Ferrari},
  journal={2018 IEEE/CVF Conference on Computer Vision and Pattern Recognition},
  year={2016},
  pages={1209-1218},
  url={https://api.semanticscholar.org/CorpusID:4396518}
}

@inproceedings{dino,
  title={Emerging properties in self-supervised vision transformers},
  author={Caron, Mathilde and Touvron, Hugo and Misra, Ishan and J{\'e}gou, Herv{\'e} and Mairal, Julien and Bojanowski, Piotr and Joulin, Armand},
  booktitle={Proceedings of the IEEE/CVF international conference on computer vision},
  pages={9650--9660},
  year={2021}
}

@inproceedings{clip,
  title={Learning transferable visual models from natural language supervision},
  author={Radford, Alec and Kim, Jong Wook and Hallacy, Chris and Ramesh, Aditya and Goh, Gabriel and Agarwal, Sandhini and Sastry, Girish and Askell, Amanda and Mishkin, Pamela and Clark, Jack and others},
  booktitle={International conference on machine learning},
  pages={8748--8763},
  year={2021},
  organization={PmLR}
}

@inproceedings{siglip,
  title={Sigmoid loss for language image pre-training},
  author={Zhai, Xiaohua and Mustafa, Basil and Kolesnikov, Alexander and Beyer, Lucas},
  booktitle={Proceedings of the IEEE/CVF international conference on computer vision},
  pages={11975--11986},
  year={2023}
}

@article{siglip2,
  title={Siglip 2: Multilingual vision-language encoders with improved semantic understanding, localization, and dense features},
  author={Tschannen, Michael and Gritsenko, Alexey and Wang, Xiao and Naeem, Muhammad Ferjad and Alabdulmohsin, Ibrahim and Parthasarathy, Nikhil and Evans, Talfan and Beyer, Lucas and Xia, Ye and Mustafa, Basil and others},
  journal={arXiv preprint arXiv:2502.14786},
  year={2025}
}

@article{tulip,
  title={TULIP: Towards Unified Language-Image Pretraining},
  author={Tang, Zineng and Lian, Long and Eisape, Seun and Wang, XuDong and Herzig, Roei and Yala, Adam and Suhr, Alane and Darrell, Trevor and Chan, David M},
  journal={arXiv preprint arXiv:2503.15485},
  year={2025}
}

@article{radiov2.5,
  title={RADIO Amplified: Improved Baselines for Agglomerative Vision Foundation Models},
  author={Heinrich, Greg and Ranzinger, Mike and Lu, Yao and Kautz, Jan and Tao, Andrew and Catanzaro, Bryan and Molchanov, Pavlo and others},
  journal={arXiv preprint arXiv:2412.07679},
  year={2024}
}

@inproceedings{adcc,
  title={Revisiting the evaluation of class activation mapping for explainability: A novel metric and experimental analysis},
  author={Poppi, Samuele and Cornia, Marcella and Baraldi, Lorenzo and Cucchiara, Rita},
  booktitle={Proceedings of the IEEE/CVF Conference on Computer Vision and Pattern Recognition},
  pages={2299--2304},
  year={2021}
}

@article{capi,
  title={Cluster and Predict Latents Patches for Improved Masked Image Modeling},
  author={Darcet, Timoth{\'e}e and Baldassarre, Federico and Oquab, Maxime and Mairal, Julien and Bojanowski, Piotr},
  journal={arXiv preprint arXiv:2502.08769},
  year={2025}
}

@article{dpt,
  title={Vision Transformers for Dense Prediction},
  author={Ren{\'e} Ranftl and Alexey Bochkovskiy and Vladlen Koltun},
  journal={2021 IEEE/CVF International Conference on Computer Vision (ICCV)},
  year={2021},
  pages={12159-12168},
  url={https://api.semanticscholar.org/CorpusID:232352612}
}

@article{mask2former,
  title={Masked-attention Mask Transformer for Universal Image Segmentation},
  author={Bowen Cheng and Ishan Misra and Alexander G. Schwing and Alexander Kirillov and Rohit Girdhar},
  journal={2022 IEEE/CVF Conference on Computer Vision and Pattern Recognition (CVPR)},
  year={2021},
  pages={1280-1289},
  url={https://api.semanticscholar.org/CorpusID:244799297}
}

@inproceedings{detr,
  title={End-to-end object detection with transformers},
  author={Carion, Nicolas and Massa, Francisco and Synnaeve, Gabriel and Usunier, Nicolas and Kirillov, Alexander and Zagoruyko, Sergey},
  booktitle={European conference on computer vision},
  pages={213--229},
  year={2020},
  organization={Springer}
}

@inproceedings{depth_anything,
  title={Depth anything: Unleashing the power of large-scale unlabeled data},
  author={Yang, Lihe and Kang, Bingyi and Huang, Zilong and Xu, Xiaogang and Feng, Jiashi and Zhao, Hengshuang},
  booktitle={Proceedings of the IEEE/CVF Conference on Computer Vision and Pattern Recognition},
  pages={10371--10381},
  year={2024}
}

@inproceedings{dvt,
  title={Denoising vision transformers},
  author={Yang, Jiawei and Luo, Katie Z and Li, Jiefeng and Deng, Congyue and Guibas, Leonidas and Krishnan, Dilip and Weinberger, Kilian Q and Tian, Yonglong and Wang, Yue},
  booktitle={European Conference on Computer Vision},
  pages={453--469},
  year={2024},
  organization={Springer}
}

@misc{vitar,
      title={ViTAR: Vision Transformer with Any Resolution}, 
      author={Qihang Fan and Quanzeng You and Xiaotian Han and Yongfei Liu and Yunzhe Tao and Huaibo Huang and Ran He and Hongxia Yang},
      year={2024},
      eprint={2403.18361},
      archivePrefix={arXiv},
      primaryClass={cs.CV},
      url={https://arxiv.org/abs/2403.18361}, 
}

@inproceedings{odise,
  title={Open-vocabulary panoptic segmentation with text-to-image diffusion models},
  author={Xu, Jiarui and Liu, Sifei and Vahdat, Arash and Byeon, Wonmin and Wang, Xiaolong and De Mello, Shalini},
  booktitle={Proceedings of the IEEE/CVF Conference on Computer Vision and Pattern Recognition},
  pages={2955--2966},
  year={2023}
}

@article{flashattention,
  title={Flashattention: Fast and memory-efficient exact attention with io-awareness},
  author={Dao, Tri and Fu, Dan and Ermon, Stefano and Rudra, Atri and R{\'e}, Christopher},
  journal={Advances in neural information processing systems},
  volume={35},
  pages={16344--16359},
  year={2022}
}

@article{adamw,
  title={Decoupled weight decay regularization},
  author={Loshchilov, Ilya and Hutter, Frank},
  journal={arXiv preprint arXiv:1711.05101},
  year={2017}
}

@article{opticam,
  title={Opti-CAM: Optimizing saliency maps for interpretability},
  author={Zhang, Hanwei and Torres, Felipe and Sicre, Ronan and Avrithis, Yannis and Ayache, Stephane},
  journal={Computer Vision and Image Understanding},
  volume={248},
  pages={104101},
  year={2024},
  publisher={Elsevier}
}

@article{dino_meets_text,
  title={DINOv2 Meets Text: A Unified Framework for Image- and Pixel-Level Vision-Language Alignment},
  author={Cijo Jose and Th{\'e}o Moutakanni and Dahyun Kang and Federico Baldassarre and Timoth{\'e}e Darcet and Hu Xu and Shang-Wen Li and Marc Szafraniec and Michael Ramamonjisoa and Maxime Oquab and Oriane Sim'eoni and Huy V. Vo and Patrick Labatut and Piotr Bojanowski},
  journal={ArXiv},
  year={2024},
  volume={abs/2412.16334},
  url={https://api.semanticscholar.org/CorpusID:274981573}
}

@inproceedings{dinotracker,
  title={Dino-tracker: Taming dino for self-supervised point tracking in a single video},
  author={Tumanyan, Narek and Singer, Assaf and Bagon, Shai and Dekel, Tali},
  booktitle={European Conference on Computer Vision},
  pages={367--385},
  year={2024},
  organization={Springer}
}

@inproceedings{maskclip,
  title={Extract free dense labels from clip},
  author={Zhou, Chong and Loy, Chen Change and Dai, Bo},
  booktitle={European Conference on Computer Vision},
  pages={696--712},
  year={2022},
  organization={Springer}
}

@inproceedings{sclip,
  title={Sclip: Rethinking self-attention for dense vision-language inference},
  author={Wang, Feng and Mei, Jieru and Yuille, Alan},
  booktitle={European Conference on Computer Vision},
  pages={315--332},
  year={2024},
  organization={Springer}
}

@article{lanczos,
  title={Lanczos filtering in one and two dimensions},
  author={Duchon, Claude E},
  journal={Journal of Applied Meteorology},
  year={1979},
}

@article{cubic,
  title={Cubic spline interpolation},
  author={McKinley, Sky and Levine, Megan},
  journal={College of the Redwoods},
  year={1998}
}

@book{cardinalspline,
  title={Cardinal spline interpolation},
  author={Schoenberg, Isaac J},
  year={1973},
  publisher={SIAM}
}

@article{maeland1988comparison,
  title={On the comparison of interpolation methods},
  author={Maeland, Einar},
  journal={IEEE transactions on medical imaging},
  year={1988},
}

@inproceedings{bevformer,
  author       = {Zhiqi Li and
                  Wenhai Wang and
                  Hongyang Li and
                  Enze Xie and
                  Chonghao Sima and
                  Tong Lu and
                  Yu Qiao and
                  Jifeng Dai},
  title        = {BEVFormer: Learning Bird's-Eye-View Representation from Multi-camera
                  Images via Spatiotemporal Transformers},
  booktitle    = {Computer Vision - {ECCV} 2022},
  year         = {2022},
}

@inproceedings{pointbev,
  author       = {Lo{\"{\i}}ck Chambon and
                  {\'{E}}loi Zablocki and
                  Micka{\"{e}}l Chen and
                  Florent Bartoccioni and
                  Patrick P{\'{e}}rez and
                  Matthieu Cord},
  title        = {PointBeV: {A} Sparse Approach to BeV Predictions},
  booktitle    = {{IEEE/CVF} Conference on Computer Vision and Pattern Recognition,
                  {CVPR} 2024},
  year         = {2024},
}

@inproceedings{harley2023simple,
author = {Adam W. Harley and Zhaoyuan Fang and Jie Li and Rares Ambrus and Katerina Fragkiadaki},
title = {Simple-{BEV}: What Really Matters for Multi-Sensor BEV Perception?},
booktitle = {IEEE International Conference on Robotics and Automation (ICRA)},
year = {2023}
}

@article{rope,
  title={Roformer: Enhanced transformer with rotary position embedding},
  author={Su, Jianlin and Ahmed, Murtadha and Lu, Yu and Pan, Shengfeng and Bo, Wen and Liu, Yunfeng},
  journal={Neurocomputing},
  volume={568},
  pages={127063},
  year={2024},
  publisher={Elsevier}
}

@inproceedings{
          diffcut,
          title={DiffCut: Catalyzing Zero-Shot Semantic Segmentation with Diffusion Features and Recursive Normalized Cut},
          author={Paul Couairon and Mustafa Shukor and Jean-Emmanuel HAUGEARD and Matthieu Cord and Nicolas THOME},
          booktitle={The Thirty-eighth Annual Conference on Neural Information Processing Systems},
          year={2024},
          url={https://openreview.net/forum?id=N0xNf9Qqmc}
          }

@misc{mmcv,
    title={{MMCV: OpenMMLab} Computer Vision Foundation},
    author={MMCV Contributors},
    howpublished = {\url{https://github.com/open-mmlab/mmcv}},
    year={2018}
}

@misc{timm,
  author = {Ross Wightman},
  title = {PyTorch Image Models},
  year = {2019},
  publisher = {GitHub},
  journal = {GitHub repository},
  doi = {10.5281/zenodo.4414861},
  howpublished = {\url{https://github.com/rwightman/pytorch-image-models}}
}

@misc{loftup,
      title={LoftUp: Learning a Coordinate-Based Feature Upsampler for Vision Foundation Models}, 
      author={Haiwen Huang and Anpei Chen and Volodymyr Havrylov and Andreas Geiger and Dan Zhang},
      year={2025},
      eprint={2504.14032},
      archivePrefix={arXiv},
      primaryClass={cs.CV},
      url={https://arxiv.org/abs/2504.14032}, 
}

@inproceedings{sam,
  title={Segment anything},
  author={Kirillov, Alexander and Mintun, Eric and Ravi, Nikhila and Mao, Hanzi and Rolland, Chloe and Gustafson, Laura and Xiao, Tete and Whitehead, Spencer and Berg, Alexander C and Lo, Wan-Yen and others},
  booktitle={Proceedings of the IEEE/CVF international conference on computer vision},
  pages={4015--4026},
  year={2023}
}

@inproceedings{swin,
  title={Swin transformer: Hierarchical vision transformer using shifted windows},
  author={Liu, Ze and Lin, Yutong and Cao, Yue and Hu, Han and Wei, Yixuan and Zhang, Zheng and Lin, Stephen and Guo, Baining},
  booktitle={Proceedings of the IEEE/CVF international conference on computer vision},
  pages={10012--10022},
  year={2021}
}

@article{hassani2025generalized,
  title={Generalized Neighborhood Attention: Multi-dimensional Sparse Attention at the Speed of Light},
  author={Hassani, Ali and Zhou, Fengzhe and Kane, Aditya and Huang, Jiannan and Chen, Chieh-Yun and Shi, Min and Walton, Steven and Hoehnerbach, Markus and Thakkar, Vijay and Isaev, Michael and others},
  journal={arXiv preprint arXiv:2504.16922},
  year={2025}
}

@inproceedings{pan2023slide,
  title={Slide-transformer: Hierarchical vision transformer with local self-attention},
  author={Pan, Xuran and Ye, Tianzhu and Xia, Zhuofan and Song, Shiji and Huang, Gao},
  booktitle={Proceedings of the IEEE/CVF conference on computer vision and pattern recognition},
  pages={2082--2091},
  year={2023}
}

@INPROCEEDINGS{nuscenes,
  title={nuScenes: A multimodal dataset for autonomous driving},
  author={Holger Caesar and Varun Bankiti and Alex H. Lang and Sourabh Vora and 
          Venice Erin Liong and Qiang Xu and Anush Krishnan and Yu Pan and 
          Giancarlo Baldan and Oscar Beijbom}, 
  booktitle={CVPR},
  year=2020
}

\newpage
\clearpage
\appendix

\title{JAFAR: Jack up Any Feature at Any Resolution \\ \normalfont{Supplementary Material}}
\maketitleappendix

\section{Additional Details}
\label{section:baselines_details}








\subsection{Evaluation}
\label{section:cam_evaluation}

To evaluate Class Activation Maps (CAMs), we employ a frozen pre-trained ViT-B/16 model as the backbone and extract Grad-CAMs. We randomly sample 2,000 images from the ImageNet validation set for which the model produces correct predictions. For each image, we compute the gradients with respect to the predicted class, average them, and use the result to weight the corresponding activation maps. The weighted activations are then summed to produce the final CAM. These activation maps are upsampled from $14 \times 14$ to $224 \times 224$, resulting in high-resolution CAMs. For each CAM, we generate a masked version of the input image by applying a binary mask that highlights regions positively associated with the model’s prediction. Formally, a masked image is obtained as $x_{\text{masked}} = x \odot \mathbbm{1}_{\text{CAM}_c(x) > 0}$. These masked images are then used to compute the evaluation metrics.

\paragraph{Average Drop} Average Drop (A.D) quantifies how much the model’s confidence in the predicted class decreases when it is presented with the masked image instead of the full image. For a single image, the metric is defined as:
\begin{equation}
    \textbf{A.D} = \frac{1}{N} \underset{i=1}{\overset{N}{\sum}}\;\frac{max(0, Y_i^c - O_i^c)}{Y_i^c} \cdot 100,
\end{equation}
where $Y_i^c$ denotes the model’s output score for class $c$ when using the full image, and $O_i^c$ denotes the score when using the masked version derived from the explanation map. The final A.D value is computed by averaging over a set of $N$  images.

\paragraph{Average Increase} Average Increase (A.I) measures how often the model’s confidence in the predicted class is higher when using the masked image than when using the full image. It is defined as:
\begin{equation}
    \textbf{A.I} = 100 \cdot \underset{i=1}{\overset{N}{\sum}}\;\frac{\mathbbm{1}_{Y_i^c < O_i^c}}{N},
\end{equation}
where $Y_i^c$ is the model’s output score for class $c$ when using the full image, and $O_i^c$ is the score when using the masked image based on the explanation map. The metric reflects the percentage of images where the explanation-based input yields a higher confidence score than the original image.

\paragraph{Average Gain \cite{opticam}} Average Gain (A.G) quantifies the improvement in predictive confidence for the target class when using the masked image instead of the full image. It is defined as:
\begin{equation}
    \textbf{A.G} = \frac{1}{N}\underset{i=1}{\overset{N}{\sum}}\;\frac{max(0, O_i^c - Y_i^c)}{1 - Y_i^c} \cdot 100,
\end{equation}
where $Y_i^c$ is the model’s output score for class $c$ on the full image, and $O_i^c$ is the score when using the masked version derived from the explanation map. This metric captures how much the explanation enhances the model’s confidence, normalized by the room for improvement $(1 - Y_i^c)$.

\paragraph{Coherency \cite{adcc}} A Class Activation Map should highlight all the relevant features that contribute to a model’s prediction, while suppressing irrelevant ones in a coherent and consistent manner. Consequently, for a given input image $x$ and a target class $c$, the CAM should remain unchanged when the image is conditioned on the CAM itself. This self-consistency can be expressed as:
\begin{equation}
    \text{CAM}_c(x \odot \text{CAM}_c(x)) = \text{CAM}_c(x)
\end{equation}
where $\odot$ denotes element-wise multiplication. This condition implies that the CAM produced from the masked image should be identical to the original CAM, ensuring that the explanation is stable. Following the approach in \cite{adcc}, we use the Pearson Correlation Coefficient between the original CAM and the CAM obtained after masking:
\begin{equation}
    \text{Coherency}(x) = \frac{\text{Cov}(\text{CAM}_c(x \odot \text{CAM}_c(x)), \text{CAM}_c(x))}{\sigma_{\text{CAM}_c(x \odot \text{CAM}_c(x))} \sigma_{\text{CAM}_c(x))}}
\end{equation}
where $\text{Cov}$ denotes the covariance and $\sigma$ the standard deviation of each CAM. Since the Pearson Correlation Coefficient ranges from $-1$ to $1$, we normalize it to the range $[0, 1]$ and express it as a percentage for interpretability. A coherency score of 100\% indicates that the attribution method is fully invariant to input perturbations guided by its own explanations.

\paragraph{Complexity}

In addition to ensuring that a CAM is coherent—preserving predictive features while discarding irrelevant ones—it is also desirable for the CAM to be as simple as possible. That is, it should highlight the minimal subset of pixels necessary to explain the model’s prediction. To quantify this notion of simplicity, we use the $\ell_0$ norm as a proxy for the Complexity of a CAM:
\begin{equation}
    \text{Complexity}(x) = \lVert \text{CAM}_c(x) \rVert_0,
\end{equation}
where $\lVert \cdot \rVert_0$ counts the number of non-zero (i.e., activated) pixels. A lower Complexity score indicates that the attribution method focuses on fewer, more relevant regions, thereby producing more concise and interpretable explanations.

\paragraph{ADCC \cite{adcc}} Since each individual metric captures a distinct aspect of CAM quality, we compute an aggregated evaluation metric—Average DCC (ADCC)—which combines Coherency, Complexity, and Average Drop into a single score using the harmonic mean:
\begin{equation}
    \text{ADCC}(x) = 3 \left( \frac{1}{\text{Coherency}(x)} + \frac{1}{1 - \text{Complexity}(x)} + \frac{1}{1 - \text{A.D}(x)}\right)^{-1}
\end{equation}
ADCC offers a unified, single-valued measure that enables direct and consistent comparison. By balancing coherency, sparsity (via low complexity), and confidence preservation (via low Average Drop), it provides a more comprehensive assessment of attribution quality.

\subsection{Baselines}
\begin{itemize}
    \item \textbf{Large Image}: For the Large Image baseline, we upsampled the original image via bilinear upsampling and use it as input to the foundation vision encoder. During evaluation on downstream tasks (see \cref{table:probing_dinov2_supervised,table:cam_table}), we upsample the input to the maximum ratio that fits in memory (i.e., $\times 8$), and subsequently apply bilinear upsampling to the resulting feature map to match the target output resolution. Due to the high computational cost and training time, we omit results on the COCO dataset. For efficiency, on Open-Vocabulary segmentation we limit upsampling to a $\times 2$ ratio in \cref{tab:maskclip_upsampling_all_column}.
    
    \item \textbf{Strided}: To obtain higher-resolution feature maps, we modify the stride of the ViT backbone to produce more patches. While the stride typically equals the patch size (e.g., 14 in DINOv2), we reduce the former to 6 in our experiments corresponding to a $\times 2.3$ upsampling. We then apply bilinear upsampling to the resulting feature map to reach the desired output resolution.
    
    \item \textbf{CARAFE} \cite{carafe}: For CARAFE, we use the CARAFEPack module from MMCV \cite{mmcv}, stacking four upsampling stages with a $\times 2$ ratio each, resulting in a final feature map upsampled by a factor of $\times 16$.
    
    \item \textbf{SAPA} \cite{sapa}: We adopt the default implementation from the official \href{https://github.com/poppinace/sapa/blob/main/SAPA.py}{SAPA repository}, stacking four SAPA upsampling modules, each with an upsampling factor of 2. This results in a final feature map with a total upsampling factor of $\times 16$.

    \item \textbf{DySample} \cite{dysample}:  We adopt the default implementation from the official \href{https://github.com/tiny-smart/dysample/blob/main/dysample.py}{Dysample repository}, stacking four Dysample upsampling modules, each with an upsampling factor of 2. This results in a final feature map with a total upsampling factor of $\times 16$.

    \item \textbf{ReSFU} \cite{resfu}: We use the default implementation from the official \href{https://github.com/zmhhmz/ReSFU/blob/master/demo/resfu.py}{ReSFU repository}, performing a direct upsampling to the target output resolution.

    \item \textbf{FeatUp} \cite{featup}: For FeatUp, we use the scalable JBU variant from the official \href{https://github.com/mhamilton723/FeatUp/blob/main/featup/upsamplers.py}{FeatUp respository} stacking 4 upsampling modules to achieve a total upsampling factor of $\times 16$. We re-train FeatUp using the provided training scripts. While the original paper trains FeatUp on the COCO dataset for 2,000 steps with a batch size of 4 and evaluates on the same dataset, we train it on the ImageNet training set for 50,000 steps ($25 \times$ more) using the same batch size, ensuring a fair comparison across methods.
    
    \item \textbf{LiFT} \cite{lift}: For LiFT, we slightly adapt the official \href{https://github.com/saksham-s/lift/blob/main/lift.py}{implementation} by resizing the intermediate representations after the downsampling module to ensure compatibility with backbones using a patch size of 14 (i.e., a downsampling factor of 14). In the official code, LiFT performs a $\times 2$ upsampling and then relies on a bilinear interpolation to upsample the features to the target output resolution. 
\end{itemize}

We summarize in \cref{tab:upsampling-comparison} the differences between upsamplers.

\begin{table}[h]
\centering
\caption{Comparison of feature upsampling methods.}
\scalebox{0.7}{
\makebox[0.7\textwidth]{
\begin{tabular}{lcccc}
\toprule
\textbf{Method} & \textbf{Task-Agnostic} & \textbf{Direct Upsampling} & \textbf{Lightweight Inference} \\
\midrule
CARAFE - SAPA - DySample           & \xmark & \xmark & \cmark  \\
ReSFU & \xmark & \cmark & \cmark  \\
LiFT             & \cmark & \xmark & \cmark  \\
FeatUp (JBU)     & \cmark & \xmark & \cmark  \\
FeatUp (Implicit)& \cmark & \cmark & \xmark  \\
\textbf{JAFAR}   & \cmark & \cmark & \cmark  \\
\bottomrule
\end{tabular}%
}}
\label{tab:upsampling-comparison}
\end{table}

\section{Additional Visualizations}
We provide in the following subsections additional comparison visualizations for upsampled feature maps \cref{features_viz_supp}, class activation maps predictions \cref{cam_viz_supp}, depth estimation \cref{depth_viz_supp} and semantic segmentation \cref{seg_viz_supp}.

\subsection{Feature Visualization}
\label{features_viz_supp}

\begin{figure}[!h]
\makebox[1.\textwidth]{
  \centering
  \includegraphics[scale=0.6]{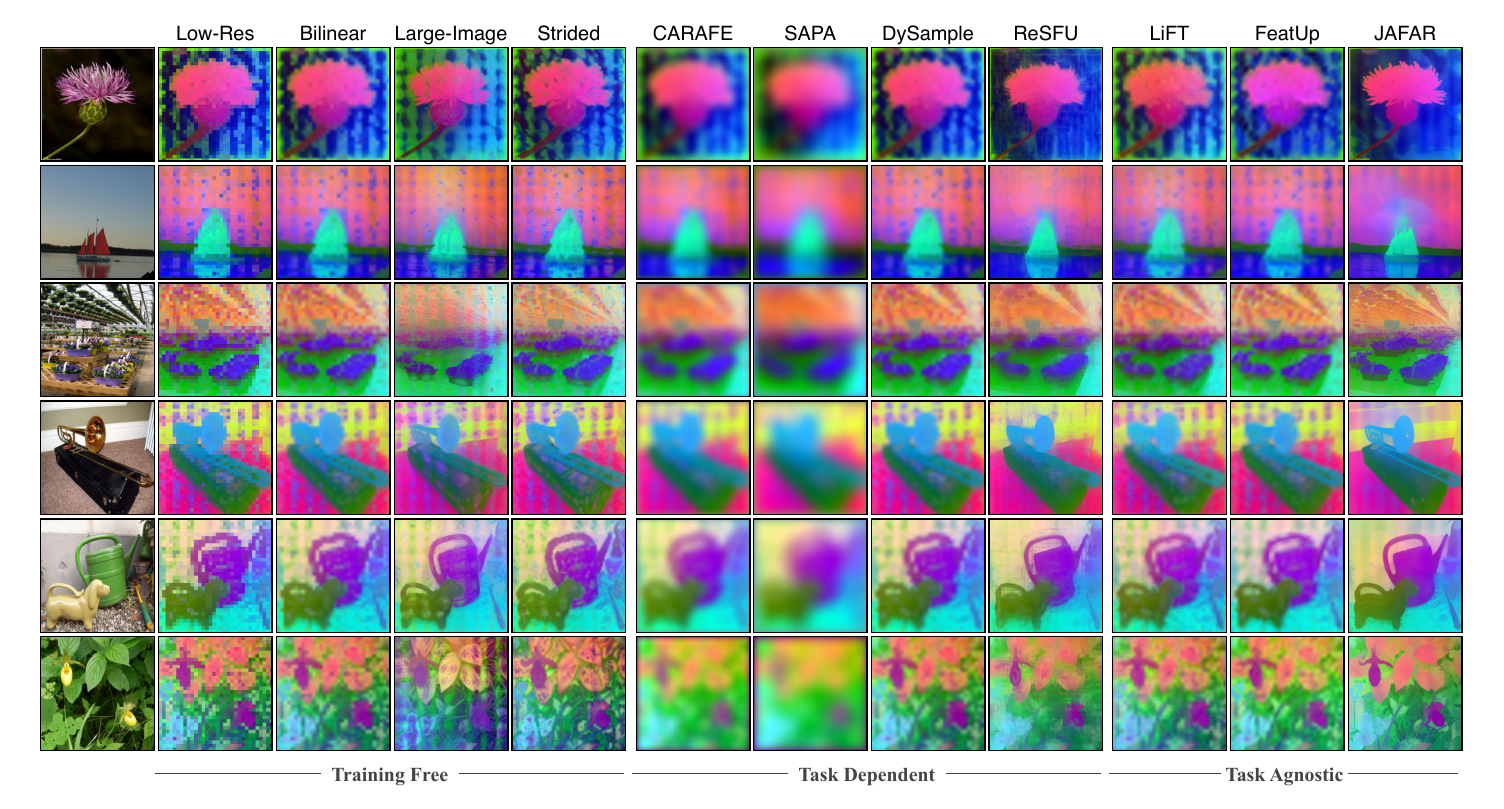}}
  \caption{\textbf{PCA Feature Visualization.} DINOv2 ViT-S/14 features at $32 \times 32$ resolution from the ImageNet validation set are upsampled to $448 \times 448$.}
  \label{fig:features_comparison_supp}
\end{figure}

\cref{fig:features_comparison_supp} shows additional PCA visualizations of upsampled feature maps. Starting from $32 \times 32$ feature maps extracted using a DINOv2-S/14 backbone, each is upsampled to $448 \times 448$ using different baseline methods. These baselines—whether training-free, task-dependent, or task-agnostic—tend to introduce varying degrees of blurriness and visual artifacts. In contrast, JAFAR, while remaining task-agnostic, produces sharp, content-aware features with minimal artifacts.

\subsection{Class Activation Maps}
\label{cam_viz_supp}

\begin{figure}[!h]
\makebox[1.\textwidth]{
  \centering
  \includegraphics[scale=0.6]{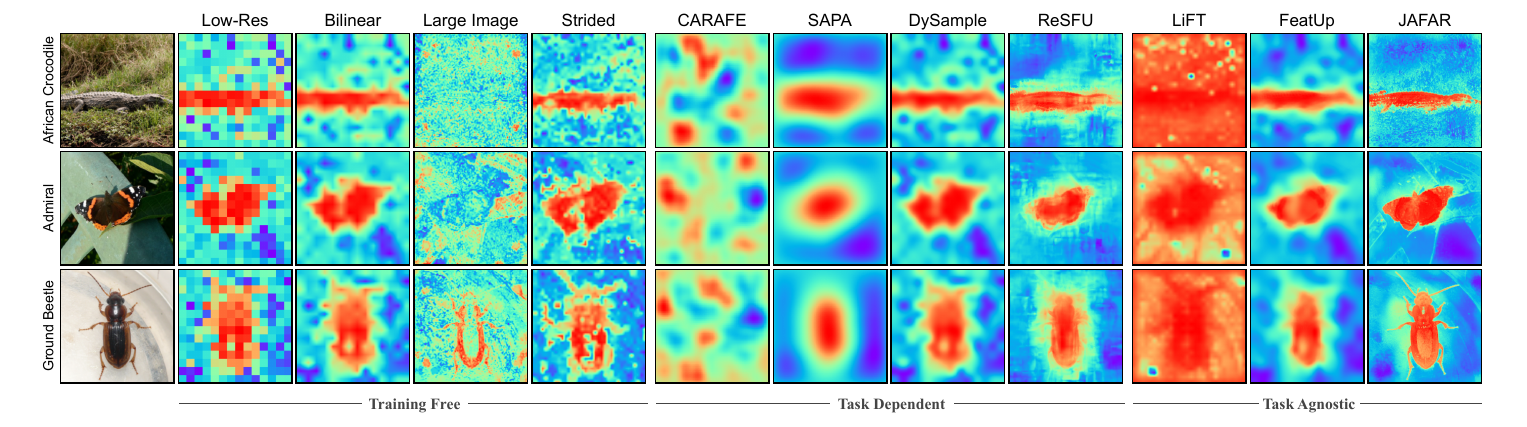}}
  \caption{\textbf{Class Activation Maps comparison.}}
  \label{fig:cam_supp}
\end{figure}

We present additional Grad-CAM visualizations based on ViT-B/16 features from the ImageNet validation set in \cref{fig:cam_supp}. Except for the “Low-Res” column, where features remain at their original 14×14 resolution, all feature maps are upsampled to 224×224 before Grad-CAM extraction. The explainability maps generated by our upsampling approach are noticeably sharper and more accurate, exhibiting fewer artifacts compared to those from alternative methods. Notably, CARAFE and LiFT fail to produce meaningful explanations in this setting, suggesting that the training of these methods does not transfer effectively to these ViT-based features.

\subsection{Depth Estimation}
\label{depth_viz_supp}

\begin{figure}[!h]
\makebox[1.\textwidth]{
  \centering
  \includegraphics[scale=0.6]{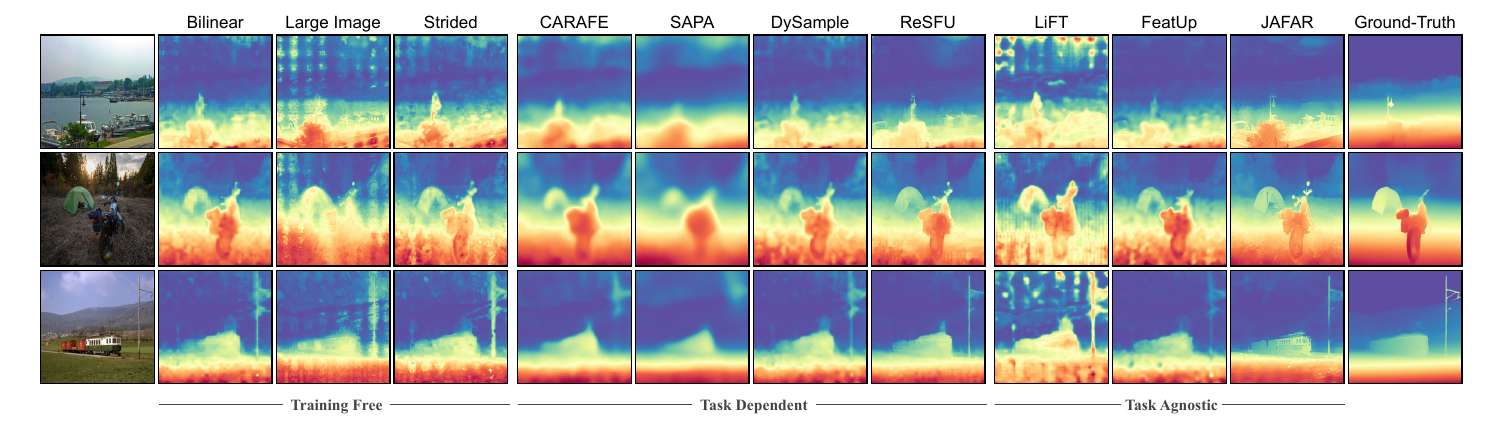}}
  \caption{\textbf{Depth Estimation Visualization.}}
  \label{fig:depth_supp}
\end{figure}

\cref{fig:depth_supp} presents additional examples of linear probe transfer learning for depth estimation on the COCO-Stuff dataset. Feature maps of size $32 \times 32$, extracted from a DINOv2-S/14 backbone, are upsampled to $448 \times 448$ using the various baseline methods. A linear probe is then trained on these features to predict depth, using supervision from a Depth-AnythingV2 model. The results demonstrate that both FeatUp variants produce high-quality features well-suited for transfer learning in depth estimation tasks.

\subsection{Semantic Segmentation}
\label{seg_viz_supp}

\begin{figure}[!ht]
\makebox[1.\textwidth]{
  \centering
  \includegraphics[scale=0.6]{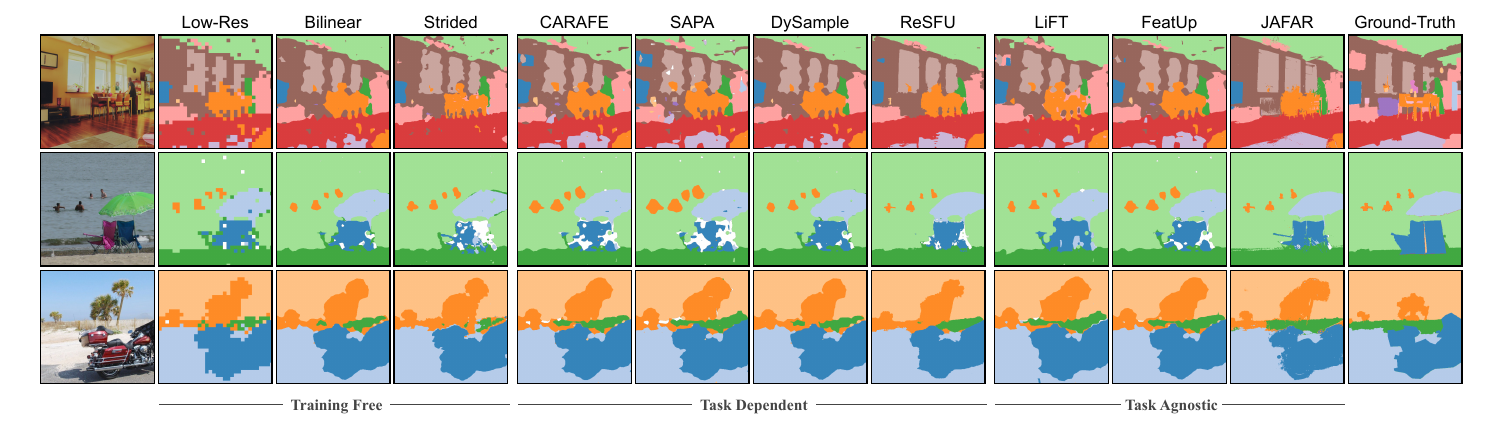}}
  \caption{\textbf{Semantic Segmentation Visualization.}}
  \label{fig:seg_supp}
\end{figure}

\cref{fig:seg_supp} presents examples of linear probe transfer learning for semantic segmentation on the COCO-Stuff dataset. Feature maps of size $32 \times 32$, extracted from a DINOv2-S/14 backbone, are upsampled to $448 \times 448$ using the various baseline methods. JAFAR produces more coherent segmentation results, offering improved delineation of both object boundaries and background regions.

\subsection{Attention Maps Visualization}

\begin{figure}[!ht]
\makebox[1.\textwidth]{
  \centering
  \includegraphics[scale=0.78]{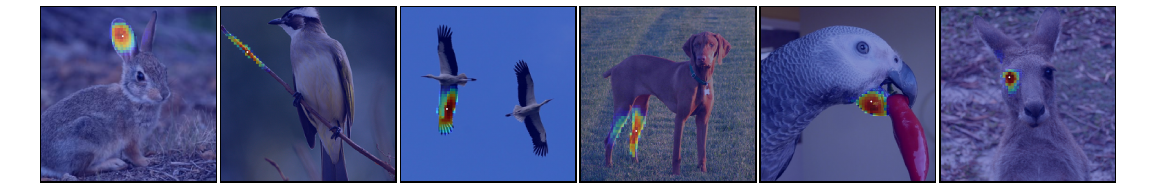}}
  \caption{\textbf{Attention maps from JAFAR.} Each map shows the attention distribution of a query location (white point) over the low-resolution keys.}
  \label{fig:attention_maps}
\end{figure}

To illustrate the behavior of the upsampling module, we visualize attention maps in \cref{fig:attention_maps}. For each example, a query location is selected, and its attention distribution over the low-resolution keys is shown. The maps reveal that attention tends to concentrate on the region semantically related to the query, even though all keys are accessible. This suggests that fully global attention may not be strictly necessary, and that localized variants could provide significant runtime and memory gains. Nevertheless, we argue that the mechanism should not be overly local, since feature maps produced by vision encoders often suffer from spatial misalignments and positional artifacts. Allowing each query a broader receptive field helps counteract these inconsistencies and improves upsampling quality. Conceptually, this resembles the principle of non-local means in image denoising, where a pixel is refined using information from a wider neighborhood rather than only its immediate surroundings.

\section{Additionnal Comparisons With Task-Agnostic Baselines}
\subsection{FeatUp \& LiFT}

In \cref{table:probing_dinov2_supervised}, we reported results obtained by training FeatUp and LiFT within our own codebase. To complement this evaluation, we present an additional comparison in \cref{tab:original_featup,tab:miou_lift_models} using the official released checkpoints, FeatUp on DINOv2 ViT-S/14 and LiFT on DINO ViT-S/16, respectively.

\begin{table}[!h]
\centering
\captionsetup{width=0.8\textwidth}
\caption{\textbf{Semantic Segmentation Comparison between JAFAR and FeatUp.} JAFAR is evaluated using FeatUp’s original feature extractor and compared against the official JBU checkpoint on the DINOv2 ViT-S/14 backbone.}
\label{tab:original_featup}
\scalebox{1.}{
\makebox[\textwidth]{
\begin{tabular}{l ccc ccc}
\toprule
 & \multicolumn{3}{c}{\textbf{224}} & \multicolumn{3}{c}{\textbf{448}} \\
\cmidrule(lr){2-4} \cmidrule(lr){5-7}
\textbf{Model} & ADE20K & Cityscapes & VOC & ADE20K & Cityscapes & VOC \\
\midrule
Bilinear & 24.50 & 38.18 & 64.10 & 28.19 & 49.35 & 70.64 \\
FeatUp   & \underline{29.26} & \underline{42.84} & \underline{71.90} & \textbf{32.87} & \underline{53.17} & \underline{77.57} \\
JAFAR    & \textbf{30.04} & \textbf{48.52} & \textbf{75.36} & \underline{32.29} & \textbf{56.45} & \textbf{79.06} \\
\bottomrule
\end{tabular}}}
\end{table}

\begin{table}[ht]
\centering
\captionsetup{width=0.8\textwidth}
\caption{\textbf{Semantic Segmentation Comparison between JAFAR and LiFT.} JAFAR is evaluated using LiFT’s original feature extractor and compared against the checkpoint on the DINO ViT-S/16 backbone.}
\label{tab:miou_lift_models}
\scalebox{0.93}{
\makebox[\textwidth]{
\begin{tabular}{l ccc ccc}
\toprule
 & \multicolumn{3}{c}{\textbf{224}} & \multicolumn{3}{c}{\textbf{448}} \\
\cmidrule(lr){2-4} \cmidrule(lr){5-7}
\textbf{Model} & ADE20K & Cityscapes & VOC & ADE20K & Cityscapes & VOC \\
\midrule
Bilinear         & 15.89 & 32.52 & 32.15 & 16.44 & 36.56 & 33.13 \\
LiFT             & \underline{16.97} & \underline{36.31} & \underline{35.30} & \underline{16.98} & \underline{40.01} & \underline{35.07} \\
LiFT-iterative   & 15.62 & 37.68 & 31.25 & 14.96 & 39.97 & 29.43 \\
JAFAR            & \textbf{21.08} & \textbf{39.93} & \textbf{47.36} & \textbf{21.44} & \textbf{43.93} & \textbf{48.18} \\
\bottomrule
\end{tabular}}}
\end{table}

Across all datasets and output resolutions, with the exception of ADE20K at 448 for FeatUp, JAFAR consistently delivers significant improvements in semantic segmentation performance. We also evaluate a LiFT-iterative baseline, which stacks two LiFT $\times 2$ upsamplers as described in the original LiFT paper. However, this iterative approach does not outperform the simpler method of applying a single LiFT $\times 2$ upsampler followed by bilinear interpolation.

\subsection{LoftUp}

We present an additional comparison with LoftUp \cite{loftup}, a recent baseline which relies on segmentation masks from SAM \cite{sam} during training and employs a two-stage pipeline that includes a self-distillation phase. For a fair evaluation, we tested JAFAR within LoftUp’s official codebase using a DINOv2 ViT-S/14 backbone.

\begin{table}[!ht]
\centering
\captionsetup{width=0.85\textwidth}
\caption{\textbf{Semantic Segmentation Comparison between JAFAR and LiFT.} JAFAR is evaluated using Loftup’s original feature extractor and compared against
the checkpoint on the DINOv2 ViT-S/14 backbone.}
\label{tab:loftup_jafar_comparison}
\begin{tabular}{llcccc}
\toprule
\multirow{2}{*}{\textbf{Resolution}} &
\multirow{2}{*}{} &
\multicolumn{2}{c}{\textbf{Cityscapes}} &
\multicolumn{2}{c}{\textbf{COCO}} \\
\cmidrule(lr){3-4} \cmidrule(lr){5-6}
& & mIoU ($\uparrow$) & Acc ($\uparrow$) & mIoU ($\uparrow$) & Acc ($\uparrow$) \\
\midrule
& LoftUp & 15.30 & 75.50 & 25.33 & 54.06 \\
$56^2$ & JAFAR  & \textbf{19.09} & \textbf{79.34} & \textbf{28.79} & \textbf{57.86} \\
& JAFAR + distillation & 18.35 & 79.05 & 28.60 & 57.49 \\
\midrule
& LoftUp & 32.02 & 86.28 & 48.89 & 73.09 \\
$112^2$ & JAFAR  & \textbf{34.56} & 87.19 & 50.67 & 74.56 \\
& JAFAR + distillation & 33.63 & \textbf{87.47} & \textbf{50.76} & \textbf{74.60} \\
\midrule
& LoftUp & 50.83 & \textbf{91.55} & 59.79 & \textbf{80.04} \\
$224^2$ & JAFAR  & 51.45 & 91.25 & 59.76 & 79.93 \\
& JAFAR + distillation & \textbf{51.84} & 91.53 & \textbf{59.90} & \textbf{80.04}  \\
\midrule
& LoftUp & \textbf{62.49} & 93.69 & 62.25 & 81.43 \\
$448^2$ & JAFAR  & 61.49 & 93.46 & 62.02 & 81.30 \\
& JAFAR + distillation & 62.30 & \textbf{93.76} & \textbf{62.36} & \textbf{81.45}  \\
\bottomrule
\end{tabular}
\end{table}

As shown in \cref{tab:loftup_jafar_comparison}, JAFAR outperforms LoftUp at lower upsampling resolutions (56 and 112) and delivers comparable performance at higher resolutions (224 and 448). Different from LoftUp, JAFAR uses a simpler and more efficient single-stage strategy: it operates entirely at low resolution and does not rely on external annotations.  Nevertheless, the self-distillation mechanism introduced in LoftUp is complementary to our approach and can be seamlessly integrated into JAFAR’s pipeline. To demonstrate this, we implemented a similar distillation objective and report the results as JAFAR + distillation in \cref{tab:loftup_jafar_comparison}. While the gains are minimal at lower resolutions, this enhancement provides a clear boost at higher resolutions (224 and 448). Lastly, JAFAR is considerably more lightweight, with only 700K parameters compared to 4.3M in LoftUp.

\section{Additional Analysis}

\subsection{Upsampling Ratios}

To evaluate the robustness of JAFAR upsampling across different ratios, we measure its performance on a linear probing semantic segmentation task at multiple scales, including extreme ratios ($8^{2} \rightarrow 896^{2}$). The results indicate that JAFAR consistently sustains strong performance as the upsampling ratio grows, underscoring its ability to generalize well beyond the training range.

\begin{table}[!h]
\centering
\captionsetup{width=0.95\textwidth}
\caption{\textbf{JAFAR robustness} across upsampling scales on linear probing segmentation.}
\label{tab:original_featup}
\scalebox{1.}{
\makebox[\textwidth]{
\begin{tabular}{c c c ccccc}
\toprule
\textbf{Input Resolution} & \textbf{Dataset}  & \textbf{Upsampler} & $56^{2}$ & $112^{2}$ & $224^{2}$ & $448^{2}$ & $896^{2}$\\
\cmidrule(lr){1-1} \cmidrule(lr){2-2} \cmidrule(lr){3-3}\cmidrule(lr){4-8}
\multirow{4}{*}{$8^{2}$} & \multirow{2}{*}{Cityscapes} & Bilinear & 31.87 & 31.81 & 31.90 & 31.61 & 31.92 \\
& & JAFAR & 36.54 & 36.04 & 35.71 & 35.77 & 35.54 \\
\cmidrule(lr){2-8}
& \multirow{2}{*}{VOC} & Bilinear & 64.79 & 64.95 & 64.89 & 64.91 & 64.9 \\
& & JAFAR & 72.7 & 73.13 & 72.77 & 72.69 & 72.55 \\

\midrule

\multirow{4}{*}{$16^{2}$} & \multirow{2}{*}{Cityscapes} & Bilinear & 47.77 & 47.77 & 47.78 & 47.83 & 47.84\\
& & JAFAR & 53.18 & 53.22 & 52.35 & 51.76 & 51.36\\
\cmidrule(lr){2-8}
& \multirow{2}{*}{VOC} & Bilinear & 75.41 & 75.42 & 75.51 & 75.50 & 75.50 \\
& & JAFAR & 80.57 & 80.76 & 80.58 & 80.36 & 80.24\\

\midrule

\multirow{4}{*}{$32^{2}$} & \multirow{2}{*}{Cityscapes} & Bilinear & 59.56 & 59.53 & 59.41 & 59.54 & 59.15\\
& & JAFAR & 59.65 & 61.39 & 61.38 & 60.86 & 60.79\\
\cmidrule(lr){2-8}
& \multirow{2}{*}{VOC} & Bilinear & 80.64 & 80.72 & 80.77 & 80.83 & 80.82\\
& & JAFAR & 81.51 & 82.53 & 82.67 & 83.55 & 83.24\\

\bottomrule
\end{tabular}}}
\end{table}

\subsection{Positional Encoding}

As shown in \cref{tab:rope_ablation}, RoPE plays a critical role in maintaining consistent spatial alignment between queries and keys across varying spatial resolutions, with its benefits becoming more pronounced at higher resolutions.

\begin{table}[h!]
\centering
\caption{\textbf{Ablation of RoPE} on linear probing segmentation (mIoU).}
\label{tab:rope_ablation}
\begin{tabular}{lcccccccc}
\toprule
 & \multicolumn{4}{c}{\textbf{Cityscapes}} & \multicolumn{4}{c}{\textbf{VOC}} \\
 \cmidrule(lr){2-5} \cmidrule(lr){6-9}
\textbf{JAFAR} & $56^2$ & $112^2$ & $224^2$ & $448^2$ & $56^2$ & $112^2$ & $224^2$ & $448^2$ \\
\midrule
w/o RoPE & 18.15 & 27.06 & 34.80 & 36.09 & 35.36 & 62.38 & 67.58 & 67.22 \\
w/ RoPE  & 21.41 & 35.80 & 52.42 & 60.86 & 40.41 & 72.78 & 80.57 & 83.55 \\
\bottomrule
\end{tabular}
\end{table}

\section{Performance}

We compare in \cref{tab:params_and_speed,tab:memory} the runtime and memory usage respectively of various methods with a batch size of 1 and input resolution of 448, across multiple target resolutions. The experiments are conducted on a single A100 GPU.

\newpage

\begin{table}[h]
\centering
\captionsetup{width=0.8\textwidth}
\caption{\textbf{Runtime (ms) for different models and resolutions.}}
\label{tab:params_and_speed}
\small
\begin{tabular}{l c cccc}
\toprule
\textbf{Model} & \textbf{\# Params (M)} & {$56^2$} & {$112^2$} & {$224^2$} & {$448^2$} \\
\midrule
\multicolumn{6}{l}{\textbf{\textit{Upsamplers}}} \\
\quad FeatUp         & 0.2 & 5.7 & 8.0 & 14.9 & 64.9 \\
\quad JAFAR          & 0.7 & 4.0 & 5.7 & 16.6 & 94.0 \\
\quad LiFT           & 1.2 & 0.9 & 0.9 & 1.0 & 1.5 \\
\quad LoftUp         & 4.3 & 3.8 & 8.9 & 24.5 & 145.1 \\
\midrule
\multicolumn{6}{l}{\textbf{\textit{Other configurations}}} \\
\quad Large Image ($\times8$)   & -   & 6.2 & 34.7 & 348 & 5 558 \\
\quad Strided (1)       & -   & 11.4 & 136.0 & 2 482 & 48 090 \\
\bottomrule
\end{tabular}
\end{table}

\begin{table}[!ht]
\centering
\caption{\textbf{Memory usage (GB) for different models and resolutions.}}
\label{tab:memory}
\begin{tabular}{lc c cccc}
\toprule
\textbf{Pass} & \textbf{Model} & \textbf{\# Params (M)} & {$56^2$} & {$112^2$} & {$224^2$} & {$448^2$} \\
\midrule
\multirow{3}{*}{Forward} 
    & Bilinear & 0.0 & 0.4 & 0.5 & 0.5 & 0.8 \\
    & FeatUp    & 0.2 & 0.6 & 0.8 & 1.6 & 4.8  \\
    & JAFAR    & 0.7 & 0.6 & 0.6 & 1.1 & 7.7 \\
    & LoftUp   & 4.3 & 0.6 & 0.7 & 1.8 & 12.3 \\
\midrule
\multirow{2}{*}{Backward} 
    & FeatUp    & 0.2 & 0.7 & 0.9 & 2.1 & 7.4 \\
    & JAFAR    & 0.7 & 0.7 & 1.1 & 3.5 & 26.0 \\
    & LoftUp   & 4.3 & 0.7 & 1.3 & 4.6 & 26.5 \\
\bottomrule
\end{tabular}
\end{table}

In \cref{tab:gflops_memory}, we report GFLOPs and training memory comparisons for the BeV segmentation setup. JAFAR delivers $6\%\!-\!15\%$ higher IoU while requiring only $3\%\!-\!10\%$ more GFLOPs than the version without upsampling, underscoring the trade-off between accuracy and computation.

\begin{table}[ht]
\centering
\caption{\textbf{Comparison of upsampling methods on BeV segmentation pipelines.}}
\label{tab:gflops_memory}
\resizebox{\textwidth}{!}{%
\begin{tabular}{lcccccc}
\toprule
\textbf{Model} & \textbf{Metric} & \textbf{No Upsampling} & \textbf{Bilinear} & \textbf{FeatUp} & \textbf{JAFAR} & \textbf{LoftUp} \\
\midrule
\multirow{3}{*}{SimpleBeV \cite{harley2023simple}} 
 & Memory (MiB) & 4539 & 4981 & 5305 & 5453 & 5305 \\
 & GFLOPs               & 594.00 & 595.10 & 595.73 & 612.24 & 621.06 \\
 & mIoU                  & 31.75 & 33.67 & 33.95 & 36.59 & -- \\
\midrule
\multirow{3}{*}{PointBeV \cite{pointbev}} 
 & Memory (MiB) & 1507 & 1833 & 1961 & 2521 & 2397 \\
 & GFLOPs               & 154.30 & 154.32 & 154.90 & 171.41 & 180.25 \\
 & mIoU                  & 34.89 & 36.01 & 35.38 & 37.20 & -- \\
\midrule
\multirow{3}{*}{BeVFormer \cite{bevformer}} 
 & Memory (MiB) & 2143 & 2489 & 2639 & 3135 & 2971 \\
 & GFLOPs               & 352.40 & 354.70 & 357.72  & 376.74 & 379.82 \\
 & mIoU                  & 33.72 & 34.18 & 34.01 & 36.54 & -- \\
\bottomrule
\end{tabular}
}
\end{table}



\section{Limitations}

Although our method offers significant advantages over existing upsampling approaches, it also has certain limitations. Because our approach employs a global attention mechanism, each query attends to all keys, and the number of keys increases with the resolution of the input image processed by the foundation vision encoder. As a result, the computational and memory costs can become prohibitive when dealing with large key sets (e.g., $64\times64$ or higher). While a large receptive field is desirable, attending to every key is not strictly necessary. Alternative attention mechanisms with more localized receptive fields \cite{swin, hassani2025generalized, pan2023slide} may achieve orders-of-magnitude efficiency improvements and represent a promising direction for future work.


\end{document}